%% file: 1586.tex
\documentclass[twoside]{article}

\usepackage[accepted]{aistats2023}
\usepackage{hyperref}
\usepackage{url}
\usepackage{amssymb}
\usepackage{amsmath}
\usepackage{cleveref}
\usepackage{graphicx}
\usepackage{float} 
\usepackage{subfigure} 
\usepackage{color}
\usepackage{booktabs}
\usepackage{listings}
\usepackage{makecell}
\usepackage{multirow}
\usepackage{natbib}
\usepackage{algorithm}
\usepackage{algorithmic}
\input{math_commands.tex}

\newcommand{\etr}{{\mathcal E_{tr}}}
\newcommand{\ete}{{\mathcal E_{te}}}
\newcommand{\ecore}{\eta_{core}}
\newcommand{\espu}{\eta_{spu}}

\begin{document}

\runningtitle{Freeze then Train: Towards Provable Representation Learning under Spurious Correlations and Feature Noise}

\runningauthor{Haotian Ye, James Zou, Linjun Zhang}

\twocolumn[

\aistatstitle{Freeze then Train: Towards Provable Representation Learning under\\Spurious Correlations and Feature Noise}

\aistatsauthor{ Haotian Ye \And James Zou$^{\dagger}$ \And  Linjun Zhang$^{\dagger}$ }

\aistatsaddress{ haotianye@pku.edu.cn\\Peking University \And  jamesz@stanford.edu\\Stanford University \And linjun.zhang@rutgers.edu\\Rutgers University } ]

\begin{abstract}
\vspace{-10pt}
The existence of spurious correlations such as image backgrounds in the training environment can make empirical risk minimization (ERM) perform badly in the test environment. 
To address this problem,  Kirichenko et al. (2022) empirically found that the core features that are related to the outcome can still be learned well even with the presence of spurious correlations.
This opens a promising strategy to first train a feature learner rather than a classifier, and then perform linear probing (last layer retraining) in the test environment. 
However, a theoretical understanding of when and why this approach works is lacking. 
In this paper, we find that core features are only learned well when their associated non-realizable noise is smaller than that of spurious features, which is not necessarily true in practice. 
We provide both theories and experiments to support this finding and to illustrate the importance of non-realizable noise. 
Moreover, we propose an algorithm called Freeze then Train (FTT), that first freezes certain salient features and then trains the rest of the features using ERM.
We theoretically show that FTT preserves features that are more beneficial to test time probing.
Across two commonly used spurious correlation datasets, FTT outperforms ERM, IRM, JTT and CVaR-DRO, with substantial improvement in accuracy (by $4.5\%$) when the feature noise is large. FTT also performs better on general distribution shift benchmarks.
\vspace{-5pt}
\end{abstract}

\input{Sections/Introduction.tex}

\input{Sections/Preliminary.tex}

\input{Sections/RatioTheorem.tex}

\input{Sections/NewMethod.tex}

\input{Sections/Experiments.tex}

\input{Sections/Conclusion.tex}

\section*{Acknowledgements}
%\textcolor{red}{TBD.}
The research of Linjun Zhang is partially supported by NSF DMS-2015378. The research of
James Zou is partially supported by funding from NSF CAREER and the Sloan Fellowship. 
In addition, we sincerely thank Haowei Lin and Ruichen Li at Peking University for providing valuable suggestions on our work.

\bibliography{reference}

% If you have textual supplementary material
\appendix
\onecolumn

\input{Sections/supplement.tex}

\vfill

\end{document}

%% file: math_commands.tex
\usepackage{amsmath,amsfonts,bm}

\newtheorem{theorem}{Theorem}

\newtheorem{lemma}{Lemma}

\def\eqref#1{equation~\ref{#1}}
\def\1{\bm{1}}

\def\vb{{\bm{b}}}

\def\vq{{\bm{q}}}

\def\vu{{\bm{u}}}
\def\vv{{\bm{v}}}

\def\vx{{\bm{x}}}

\def\mA{{\bm{A}}}
\def\mB{{\bm{B}}}
\def\mC{{\bm{C}}}
\def\mD{{\bm{D}}}

\def\mH{{\bm{H}}}
\def\mI{{\bm{I}}}

\def\mM{{\bm{M}}}

\def\mQ{{\bm{Q}}}

\def\mW{{\bm{W}}}
\def\mX{{\bm{X}}}
\def\mY{{\bm{Y}}}

\def\mLambda{{\bm{\Lambda}}}
\def\mSigma{{\bm{\Sigma}}}

\DeclareMathAlphabet{\mathsfit}{\encodingdefault}{\sfdefault}{m}{sl}
\SetMathAlphabet{\mathsfit}{bold}{\encodingdefault}{\sfdefault}{bx}{n}

\def\sD{{\mathbb{D}}}
\def\sR{{\mathbb{R}}}
\def\sS{{\mathbb{S}}}
\def\sT{{\mathbb{T}}}

\newcommand{\R}{\mathbb{R}}

%% file: Sections/Introduction.tex
\section{Introduction}

\begin{figure}[t!]
\begin{center}
\includegraphics[width=0.45\textwidth]{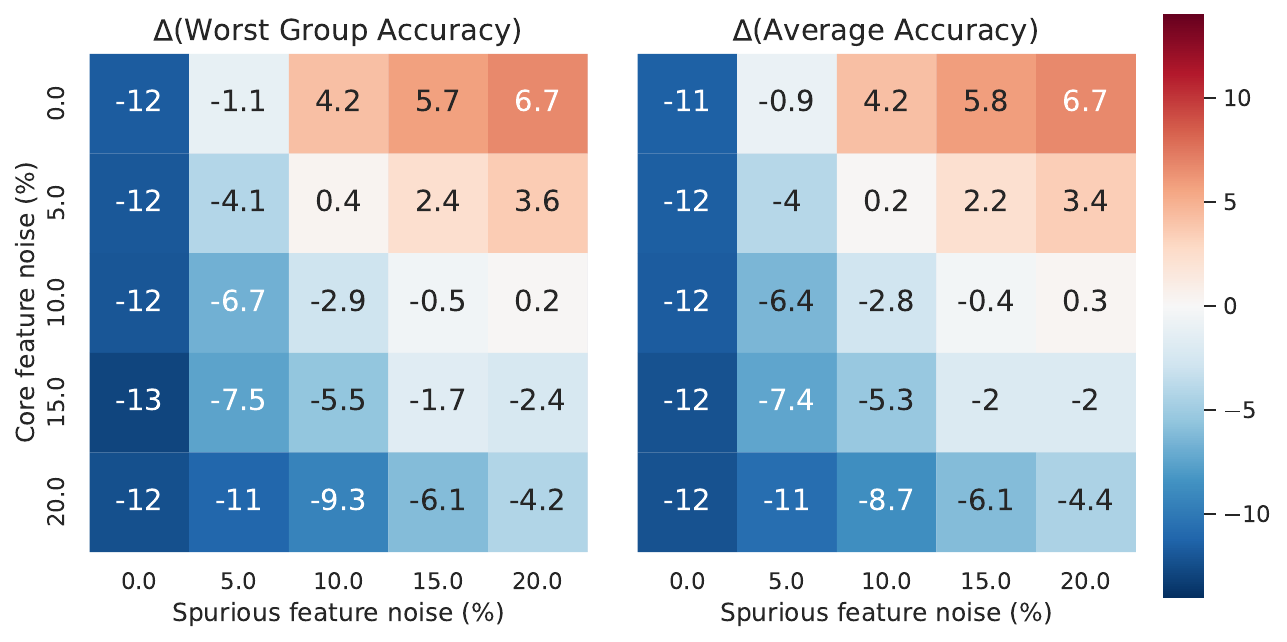}
\end{center}
\vspace{-10pt}
\caption{
{The improvement of last layer retraining accuracy (\%) \textit{before} v.s. \textit{after} ERM training on Dominoes dataset \citep{shah2020pitfalls}. 
The model is initialized with ImageNet pretrained parameters.}
The x-axis and y-axis represent  noise levels of the spurious and core features, respectively.
ERM training helps/harms the performance when the non-realizable noise of core features is smaller/greater than that of the spurious features. 
Experiment settings are in \Cref{sec:experiments}.
}
\label{fig:dominoes_erm}
\vspace{-15pt}
\end{figure}

Real-world datasets are riddled with features that are ``right for wrong reasons'' \citep{zhou2021examining}. 
% Fed with the datasets with spurious correlations, it was commonly reported that machine learning models are biased towards these spurious features.
For instance, in Waterbirds \citep{sagawa2019distributionally}, the bird type can be highly correlated with the spurious feature image backgrounds, and in CelebA \citep{liu2015deep} the hair color can be relevant to the gender.
These features are referred to as \textit{spurious features} \citep{hovy2015tagging,blodgett2016demographic,hashimoto2018fairness}, being predictive for most of the training examples, but are not truly correlated with the intrinsic labeling function.
Machine learning models that minimize the average loss on a training set (ERM) rely on these spurious features and will suffer high errors in environments where the spurious correlation changes.
Most previous works seek to avoid learning spurious features by minimizing subpopulation group loss \citep{duchi2019distributionally}, by up-weighting samples that are misclassified \citep{jtt}, by selectively mixing samples \cite{yao2022improving}, and so on.
The general goal is to recover the core features under spurious correlations.

\begin{figure*}[t]
\begin{center}
\includegraphics[width=0.9\textwidth]{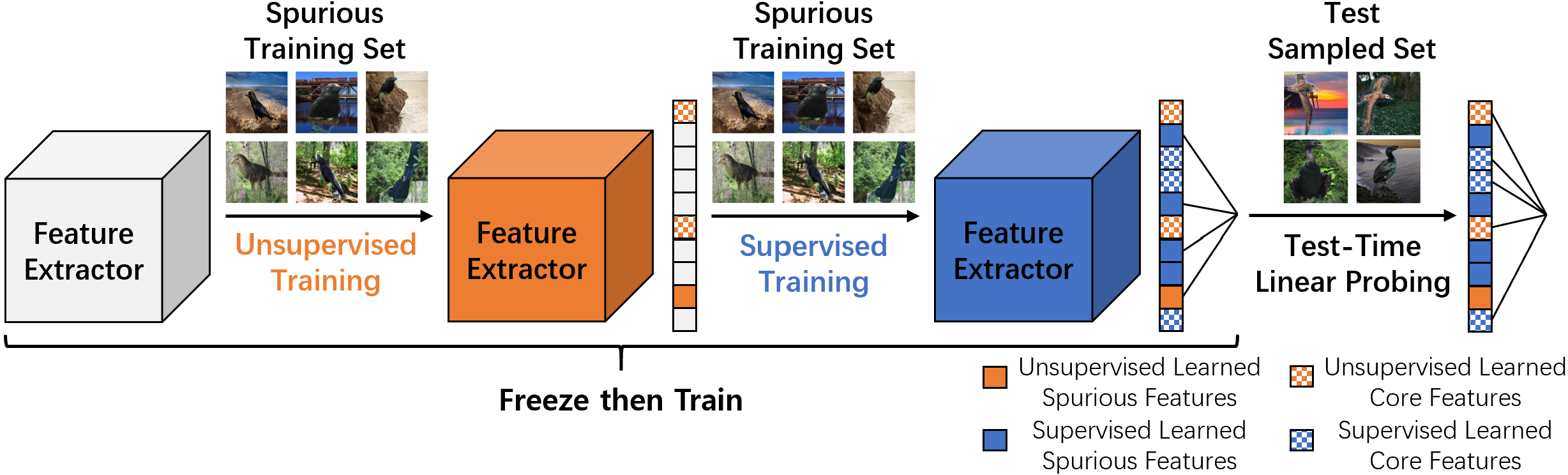}
\end{center}
\vspace{-10pt}
\caption{
An illustration of our method, Freeze then Train (FTT).
We start with a pretrained feature extractor (e.g. CNN) and find dataset-specific salient features using any unsupervised method like contrastive learning or PCA (the orange part).
We then freeze these features and learn the rest of the features using any supervised method like ERM or a robust training algorithm (the blue part).
In the test environment, the last layer is retrained.
The pseudo-code can be found in Appendix~\ref{appendix:alg}. 
% \james{Confusing that the orange and blue features are dispersed. Might be clearer to just have the top block of features be orange and bottom blue.}
}
\label{fig:pipeline}
\vspace{-15pt}
\end{figure*}

Recently, \cite{lastlayer} empirically found that ERM can still learn the core features well even with the presence of spurious correlations.
They show that by simply retraining the last layer using a small set of data with little spurious correlation, one can reweight on core features and achieves state-of-the-art performance on popular benchmark datasets.
This method is called Deep Feature Reweighting (DFR), and it points to a new promising strategy to overcome spurious correlation: learn a feature extractor rather than a classifier, and then perform linear probing on the test environment data.
This strategy is also used in many real-world applications in NLP, where the pipeline is to learn a large pretrained model and conduct linear probing in downstream tasks \citep{brown2020language}. 
It simply requires a CPU-based logistic regression on a few amount of samples from the deployed environment.

However, several problems regarding this strategy remain open. First, 
it is unclear \textit{when and why the core features can and cannot be learned during training and be recovered in test-time probing.} Moreover, in the setting where the DFR strategy does not work well, \textit{is there an alternative strategy to learn the core features and make the test-time probing strategy work again?} 

In this paper, we first present a theoretical framework to quantify this phenomenon in a two-layer linear network and give both upper and lower control of the probing accuracy in \Cref{theorem_upper,theorem_lower}.
Our theories analyze the effect of training and retraining, which is highly nontrivial due to the {non-convex} nature of the problem. 
{
Our theories point out an essential factor of this strategy: \textbf{the feature-dependent non-realizable noise} (abbreviated as non-realizable noise).
Noise is common and inevitable in real-world \citep{frenay2013classification}.
For example, labels can have intrinsic variance and are imperfect, and human experts may also assign incorrect labels; in addition, noise is often heterogeneous and feature-dependent \cite{zhang2021learning}, and spurious features can be better correlated with labels in the training environment \citep{yan2014learning, veit2017learning}.
}

Our theories show that in order to learn core features, ERM requires the non-realizable noise of core features to be much smaller than that of spurious features. 
As illustrated in \Cref{fig:dominoes_erm}, when this condition is violated, the features learned by ERM perform even worse than the pretrained features. 
The intuition is that models typically learn a mixture of different features, where the proportion depends on the trade-off between information and noise: features with larger noise are used less.
During the last-layer probing, when the proportion of the core feature is small, we suffer more to amplify this feature. 
Our theories and experiments suggest that the scenario in \cite{lastlayer} is incomplete, and the strategy can sometimes be ineffective.

Inspired by this understanding, we propose an algorithm, called Freeze then Train (FTT), which first learns salient features \textit{in an unsupervised way} and freezes them, and then trains the rest of the features via \textit{supervised learning}.
We illustrate it in \Cref{fig:pipeline}.
Based on our finding that linear probing fails when the non-realizable noise of spurious features is smaller (since labels incentivize ERM to focus more on features with smaller noise), we propose to learn features both \textit{with} and \textit{without} the guidance of labels.
This exploits the information provided in labels, while still preserving useful features that might not be learned in supervised training.
We show in \Cref{theorem_FTT} that FTT attains near-optimal performance in our theoretical framework, providing initial proof of its effectiveness.

We conduct extensive experiments to show that:
(1) In real-world datasets the phenomenon matches our theories well.
(2) On three spurious correlation datasets, FTT outperforms other algorithms by $1.4\%, 0.3\%, 4.1\%$ on average, and $4.5\%, 0.4\%, 9\%$ at most.
(3) On more general OOD tasks such as three distribution shift datasets, FTT outperforms other OOD algorithms by $1.1\%, 0.8\%, 2.1\%$ on average.
(4) We also conduct fine-grained ablations experiments to study FTT under different unsupervised feature fractions, and a different number of learned features.

Together, we give a theoretical understanding of the probing strategy, propose FTT that is more suitable for test-time probing and outperforms existing algorithms in various benchmarks.
Even under spurious correlation and non-realizable noises, by combining ERM with unsupervised methods, we can still perform well in the test environment.

 \textbf{Related Works on rbustness to spurious correlations.} Recent works aim to develop methods that are robust to spurious correlations, including learning invariant representations~\citep{arjovsky2019invariant,guo2021out,khezeli2021invariance,koyama2020out,krueger2021out,yao2022improving}, weighting/sampling \citep{shimodaira2000improving, japkowicz2002class,buda2018systematic, cui2019class, sagawa2020investigation}, and distributionally robust optimization (DRO) \citep{ben2013robust,namkoong2017variance,oren2019distributionally}. 
Rather than learn a ``one-shot model'', we take a different strategy proposed in \cite{lastlayer} that conducts regression on the test environment.

 \textbf{Related Works on representation learning.} Learning a good representation is essential for the success of deep learning models \cite{bengio2013representation}. The representation learning has been studied in the settings of autoencoders \cite{he2021masked}, transfer learning \cite{du2020few,tripuraneni2020theory, tripuraneni2021provable,deng2021adversarial,yao2021meta,yang2022nearly}, topic modeling \cite{arora2016latent,ke2022using, wu2022sparse}, algorithmic fairness \cite{zemel2013learning,madras2018learning, burhanpurkar2021scaffolding} and self-supervised learning \cite{lee2020predicting,ji2021power,tian2021understanding,nakada2023understanding}.

%% file: Sections/Preliminary.tex
\section{Preliminary}
Throughout the paper, we consider the classification task $\mathcal X \to \mathcal Y$, where $\mathcal X \subset \mathbb R^d$ and $\mathcal Y = [K]$. Here we use $[N]$ to denote the set $\{1,\cdots, N\}$.
We denote all possible distributions over a set $E$ as $\Delta(E)$.
Assume that the distribution of $(\vx, y)$ is $\mathcal E_{tr}$ in the training environment and $\mathcal E_{te}$ in the test environment.

\textbf{Spurious correlation.} 
Learning under spurious correlation is a special kind of Out-Of-Distribution (OOD) learning where $\sD_{tr} \not= \sD_{te}$.
{We denote the term feature as a mapping $\phi(\cdot):\mathcal X \mapsto \sR^m$ that captures some property of $\mathcal X$.}
We say $\phi$ is \textbf{core (robust)} if $y\mid\phi(x)$ has the same distribution across $\mathcal E_{tr}$ and $\mathcal E_{te}$.  
Otherwise, it is \textbf{spurious}.

\textbf{Non-realizable noise.}
Learning under noise has been widely explored in machine learning literature, but is barely considered when spurious correlations exist.
Following \cite{buhlmann2020invariance, arjovsky2019invariant}, we consider non-realizable noise as the randomness along a generating process (can be either on features or on labels).
Specifically, in the causal path $\phi(x)_{core}\to y$, we treat the label noise on $y$ as the non-realizable noise, and call it ``core noise'' as it is relevant to the core features; in the causal path $y\to  \phi_{spu}(x)$, we treat the feature noise on $\phi_{spu}(x)$ as the non-realizable noise, and call it ``spurious noise'' as it is relevant to the spurious features.
As we will show, the non-realizable noise influences the model learning preference.

\textbf{Goal.}
Our goal is to minimize the prediction error in $\ete$, where spurious correlations are different from $\etr$.
In this paper, we consider the new strategy proposed in \cite{lastlayer} that trains a feature learner on $\etr$ and linearly probes the learned features on $\ete$, which we call \textbf{test-time probing} (or \textbf{last layer retraining}).
No knowledge about $\ete$ is obtained during the first training stage.
When deploying the model to $\ete$, we are given a small test datasets $\{\vx_i, y_i\}_{i=1}^n$ sampled from $\sD_{te}$, and we are allowed to conduct logistic/linear regression on $\phi(\vx)$ and $y$ to obtain our final prediction function.
The goal is that after probing on the learned features, the model can perform well in $\ete$, \textit{under various possible feature noise settings.}

%% file: Sections/RatioTheorem.tex
\section{Theory: Understand Learned Features under Spurious Correlation}

In this section, we theoretically show why core features can still be learned by ERM in spite of spurious correlations, and why non-realizable noises are crucial.
Roughly speaking, only when core noise is smaller than spurious noise, features learned by ERM can guarantee the downstream probing performance.
All proofs are in Appendix~\ref{appendix:proofs}.

\subsection{Problem Setup}
\textbf{Data generation mechanism.}
To capture the spurious correlations and non-realizable noises, we assume the data $\big(\vx, y\big)$ is generated from the following mechanism:
\begin{align*}
&\vx_1 \sim \mathbb P \in \Delta(\R^{1 \times d_1}) 
, y = \vx_1 \beta + \epsilon_{core}, \\
&\vx_2 = \begin{cases}
y\gamma^\top + \epsilon_{spu}  & \etr \\
\epsilon_{spu} & \ete 
\end{cases}
\in \R^{1 \times d_2}  , \vx = (\vx_1,\vx_2) \in \R^{1 \times d}.
\end{align*}
Here $\vx_1$ is the core feature with an invertible covariance matrix $\mSigma \triangleq \mathbb E[x_1^\top x_1]$.
$\vx_2$ is the spurious feature that is differently distributed in $\etr$ and $\ete$.
$\epsilon_{core} \in \R, \epsilon_{spu} \in \R^{1 \times d_2}$ are independent core and spurious noises with mean zero and variance (covariance matrix) $\eta_{core}^2$ and $\eta_{spu}^2 \mI$ respectively. 
{
$\beta \in \R^{d_1 \times 1}, \gamma \in \R^{d_2 \times 1}$ are normalized coefficients with unit $\ell_2$ norm.
We assume that there exists some $k\in\mathbb N$ such that the top-$k$ eigenvalues are larger than the noise variance $\eta_{spu}^2, \eta_{core}^2$, and $\beta$ lies in the span of top-$k$ eigenvectors of $\mSigma$.
This is to ensure that the signal along $\beta$ is salient enough to be learned.
For technical simplicity, we also assume that all eigenvalues of $\mSigma$ are distinct.}

Our data generation mechanism is motivated by \cite{arjovsky2019invariant} (Figure 3), where we extend their data model.
We allow core features to be drawn from any distribution $\mathbb P$ so long as $\mSigma$ is invertible, while \cite{arjovsky2019invariant} only consider a specific form of $\mathbb P$.
In addition, in our mechanism, labels depend on core features and spurious features depend on labels. 
However, our theorems and algorithms can be easily applied to another setting where both core and spurious features depend on labels. 
This is because the difference between the two settings can be summarized as the difference on $\mSigma$, while the techniques we use do not rely on the concrete form of $\mSigma$.

\textbf{Models.}
To capture the property of features and retraining, we consider a regression task using a two-layer linear network $f(\vx) = \vx\mW \vb$, where $\mW \in \R^{d\times m}$ is the feature learner and $\vb \in \R^{m \times 1}$ is the last layer that will be retrained in $\ete$. 
We assume that the model learns a low-dimensional representation ($m \ll d$), but is able to capture the ground truth signal ($m \gg k$).
Notice that the optimization over $(\mW,\vb)$ is \textit{non-convex}, and there is no closed-form solution. This two-layer network model has been commonly used in machine learning theory literature \citep{arora2018optimization,gidel2019implicit,kumar2022fine}.
The major technical difficulty 
 in our setting 
is how to analyze the learned features and control probing performance under this non-convexity with spurious correlations.
We assume the parameters are initialized according to Xavier uniform distribution\footnote{Our theorems can be easily applied to various initializations.}.

\textbf{Optimization.}
During the training stage, we minimize the $l_2$-loss $\ell_{tr}(\mW, \vb) = \frac 1 {2n} \|f(\mX)  - \mY \|^2$ where $\mX = (\vx_1^\top,\cdots,\vx_n^\top )^\top$ and $\mY = (y_1,\cdots, y_n)^\top$. 
For the clarity of analysis, we consider two extremes that can help simplify the optimization while still maintaining our key intuition.
First, we take an infinitely small learning rate such that the optimization process becomes a gradient flow \citep{gunasekar2017implicit,du2018algorithmic}. 
Denote the parameters $\mW, \vb$ at training time $t$ as $\mW(t), \vb(t)$, and $\vv(t) = \mW(t) \vb(t)$.
Second, we consider the infinite data setting ($n \to \infty$). 
This is a widely used simplification to avoid the influence of sample randomness \cite{kim2019multiaccuracy,ghorbani2021linearized}.
The parameters are updated as 
% \begin{small}
\begin{align*}
\partial_t \mW(t) &= - \nabla _ \mW \ell_{tr}(\mW(t), \vb(t))\\
&=- \left (\frac{\mX^\top \mX}{n}\mW(t)\vb(t) - \frac{\mX^\top \mY}{n}\right)\vb(t)^\top \\
&=  \left( \mathbb E[\vx^\top y]- \mathbb E [\vx^\top \vx]\vv(t) \right)\vb(t)^\top 
\\
\partial_t \vb(t) &= -\nabla_\vb \ell_{tr}(\mW(t),\vb(t)) \\
&= -\mW(t)^\top  \left (\frac{\mX^\top \mX}{n}\mW(t)\vb(t) - \frac{\mX^\top \mY}{n}\right) \\
&=  \mW(t)^\top \left( \mathbb E[\vx^\top y]- \mathbb E [\vx^\top \vx]\vv(t) \right).
\end{align*}
% \end{small}
In the test stage, we retrain the last layer $\vb$ to minimize the test loss, i.e.
% \begin{small}
$$
\ell_{te}(\mW) = \min_\vb \mathbb E_{\mathcal E_{te}} \frac 1 2 \|\vx \mW \vb - y \|^2.
$$
% \end{small}
In the test stage, the spurious correlation is broken, i.e. $\vx_2 = \epsilon_{spu}$. The minimum error in the test stage is $\text{err}^*_{te} = \eta_{core}^2/2$ when $\vv = (\beta^\top, \mathbf 0)^\top$.

\subsection{Theoretical Analysis: Noises Matter}
We are now ready to introduce our theoretical results when the core features can and cannot be learned by ERM with different levels of non-realizable noises.
One important intuition on why core features can still be learned well despite the (possibly more easily learned) spurious features is that, the loss can be further reduced by using both core and spurious features simultaneously. 
% The following lemma specify this intuition.
\begin{lemma}
    \label{lemma_optimal_training_error} For all $\mW \in \R^{d\times m}, \vb \in \R^{m}$, we have
    % \begin{small}
    $$
    \ell_{tr}(\mW, \vb) \geq \frac 1 2 \mathbb E \|\vx \vv^*_{tr} - y \|_2^2 =  \frac{\eta_{core}^2 \eta_{spu}^2}{2(\eta_{core}^2 + \eta_{spu}^2)} \triangleq {\rm err}_{tr}^*, 
    $$
    % \end{small}
    where $\vv^*_{tr} = (\alpha\beta^\top,(1-\alpha)\gamma^\top)^\top$ is the optimal coefficient for training, and $\alpha = \frac{\eta_{spu}^2}{\eta_{core}^2 + \eta_{spu}^2}$.
\end{lemma}

\Cref{lemma_optimal_training_error} shows that by assigning $\alpha$ fraction of weight to the core feature $\beta$ and the rest to $\gamma$, the loss is minimized. 
This implies that the model will learn a mixture of both features even with large spurious correlations. 
More importantly, the magnitude of $\alpha$ will largely influence the probing performance.
{
During the test stage, $\vx_2$ become useless, and the trained $\mW(t)\vb(t)$ can only recover $\alpha$ fraction of $y$, which induces a large approximation error.
To this end, during the retraining the last layer coefficients should scale up in order to predict $y$ well.
}
Meanwhile, this also scales up the weight on $\vx_2$, which is merely a harmful noise, resulting in a \textit{trade-off} between learning accurate core features and removing spurious features.
When the core noise is small, i.e., $\alpha \approx 1$, the noise on $\vx_2$ will not be scaled up much.
The Waterbirds dataset considered in \cite{lastlayer} has $\eta_{core} = 0\% < \eta_{spu} = 5\%$, falling into this region. The following theorem tells how well the ERM with last-layer probing works in this region.

\begin{theorem}[Upper Bound]
\label{theorem_upper}
Assume that $\vv(t)$ is bounded away from $\mathbf 0$ throughout the whole optimization\footnote{This is to guarantee that our gradient flow will not fail to converge to an minimum, in which case the theorem is meaningless.}, i.e. $\|\vv(t)\|_2 > c_0 >0$. 
Then, for any $0<\eta_{core} < \eta_{spu}$, any time $t$, we have
% \begin{small}
\begin{align}\label{formula_upper_bound}
\ell_{te}(\mW(t)) \leq \left( 1 + \frac{\eta_{core}^2}{\eta_{spu}^2}\right){\rm err}^*_{te} + \mathcal O(t^{-1}).
\end{align}
% \end{small}
Here $\text{err}_{te}^* = {\eta_{core}^2}/2$ is the optimal testing error and $\mathcal O$ hides the dependency on $\eta_{core}, \eta_{tspu}, c_0$ and the initialized parameters. When $\frac{\eta_{core} }{ \eta_{spu}} \to 0$, this theorem suggests test-time probing achieves near optimal error.
\end{theorem}

\Cref{theorem_upper} gives a theoretical explanation of the last layer retraining phenomenon.
It shows that the test error after retraining can be close to $\text{err}_{te}^*$ over time.
However, this guarantee holds only when $\eta_{core} < \eta_{spu}$.
The following theorem shows that when the core features have large noise, the representation learned by ERM would produce a downgraded performance after linear probing. 

\begin{figure}[t]
\begin{center}
\includegraphics[width=0.48\textwidth]{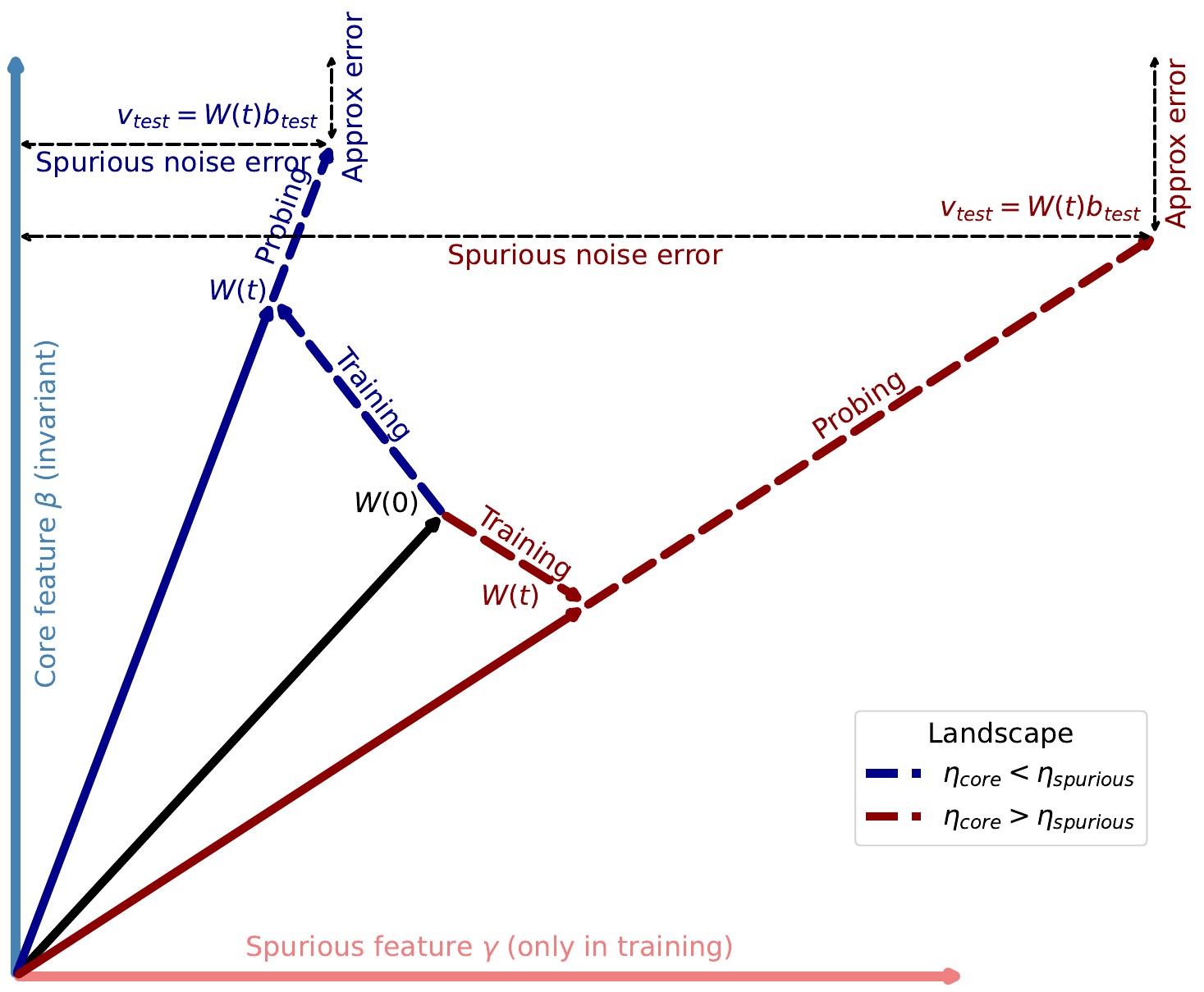}
\end{center}
\vspace{-5pt}
\caption{
A toy example illustrating when and why ERM can perform well after retraining in $\ete$ ($d = 2, m =1$). 
Assume the \textcolor[rgb]{0.275, 0.51, 0.706}{core feature $\beta$} is vertical and the \textcolor[rgb]{0.941, 0.502, 0.502}{spurious feature $\gamma$} is horizontal. 
Both features can predict $y$ in $\etr$, while $\gamma$ is useless in $\ete$ since $\vx_2 = \epsilon_{spu}$.
We initialize our single feature $\mW(0)$, and obtain $\mW(t)$ after training on $\etr$.
We then retrain the last layer (probing) on $\ete$, i.e. rescale $\mW(t)$ and obtain $\vv_{test}$.
When $\eta_{core} < \eta_{spu}$, $\mW(t)$ will use $\beta$ more (\textcolor[rgb]{	0, 0, 0.545}{the blue flow}); after probing, $\vv_{test}$ can recover $\beta$ (small approximation error) without suffering much from the spurious $\epsilon_2$ on the direction or $\gamma$ (small spurious noise error).
On the contrary, when $\eta_{core}$ is large, $\mW(t)$ will follow \textcolor[rgb]{0.545, 0, 0}{the red flow}; this leads to a trade-off between two error terms.
In this case, ERM performs much worse. 
Notice that flows in the figure are just for illustration. In practice, probing can either lengthen or shorten $\mW(t)$, depending on the concrete form of two error terms.
}
\label{fig:noise_illustration}
\vspace{-10pt}
\end{figure}

\begin{theorem}[Lower Bound]
\label{theorem_lower}
Assume that in the infinity, $\mW_1(\infty) \triangleq \lim_{t\to \infty} \mW_1(t)$ has full column rank, which almost surely holds when $m< d$. Then for any $\eta_{core} > \eta_{spu} > 0$, we have
% \begin{small}
\begin{align}
    &\quad \lim_{t\to \infty} \frac {\ell_{te}(\mW(t))}{{\rm err}^*_{te}} \notag\\
    &\geq 1+ \frac{\eta_{core}^2}{2\eta_{spu}^2} 
    \Bigg(
    % \notag\\&
    1 \wedge	\frac 1 {2\eta_{spu}^2 \left\|\mSigma^{-1}\right\|_2 \left\|\mW_1^\dagger(\infty)\right\|^2_2} \Bigg).
\end{align}
% \end{small}
Here $\mA ^\dagger$ is the Moore-Penrose inverse of $\mA$, and $a \wedge b$ takes the minimum over $a, b$. When $\frac{\eta_{spu}}{\eta_{core}} \to 0$, the last layer retraining error is much larger than the optimal error.
\end{theorem}

\Cref{theorem_lower} implies that the error can be $\frac{\eta_{core}^2}{\eta_{spu}^2}$ times larger than $\text{err}_{te}^*$ when $\eta_{core} > \eta_{spu}$, showing that ERM with last layer retraining does not work in this scenario, and the features learned by ERM are insufficient to recover near-optimal performance.
In summary, we prove that test-time probing performance largely relies on the non-realizable noises, and it only works when the core noise is relatively smaller.
We illustrate two theorems in \Cref{fig:noise_illustration}.

%% file: Sections/NewMethod.tex
\section{Method: Improving Test-Time Probing}

Our theories raise a natural question:  \textit{can we improve the learned features and make the test-time probing strategy effective under various noise conditions?}
A feature can be better correlated with labels in $\etr$ than others, but the correlation may be spurious and even disappears in $\ete$.
Without concrete knowledge about $\ete$ and spurious correlations, it is impossible to determine whether or not a learned feature is informative only in $\etr$, especially given that there are innumerable amount of features.
This problem comes from treating the label as an absolute oracle and is unlikely to be addressed by switching to other supervised robust training methods that still depend on labels. 
We experimentally verify this in \Cref{sec:main_table}.

In order to perform well in test-time probing under different noise conditions, we should also learn salient features that are selected \textit{without} relying on labels. 
This helps preserve features that are useful in the testing stage, but are ruled out because they are less informative than other features w.r.t. labels.
By learning features both with and without the help of labels, we can extract informative features and simultaneously maximize diversity. 
To this end, we propose the Freeze then Train (FTT) algorithm, which first freezes certain salient features unsupervisedly and then trains the rest of the features supervisedly.
The algorithm is illustrated in \Cref{fig:pipeline}, and we describe the details below.

\subsection{Method: Freeze then Train}

\begin{algorithm}
	%\textsl{}\setstretch{1.8}
	\renewcommand{\algorithmicrequire}{\textbf{Input:}}
	\renewcommand{\algorithmicensure}{\textbf{Output:}}
	\caption{\textbf{F}reeze \textbf{T}hen \textbf{T}rain}
	\label{alg:FTT}
	\begin{algorithmic}[1]
		\REQUIRE Dataset $\sS = \{\vx_i, y_i\}_{i=1}^n$, initialized feature extractor $\mathcal M: \mathcal X \mapsto \mathbb R^{m}$, unsupervised fraction $p$, n\_class $K$.
        \STATE Conduct PCA on $\{\mathcal M(\vx_i)\}_{i=1}^n$ with dimension $pm$, obtain transform matrix $\mW_{ul} \in \mathbb R^{m \times pm}$
        \STATE Set unsupervised model $\mathcal M_{ul}(\vx) = \mathcal M(\vx) \mW_{ul}$, and freeze its parameters (including $\mW_{ul}$) 
        \STATE set $\mathcal M_{sl}(\vx) = \mathcal M(\vx) \mW_{sl}$, initialize linear head $h: \sR^{m} \mapsto \mathbb R^K$.
        \STATE Supervisedly train the model $\mathcal M_{FTT}(\vx) = h((\mathcal M_{ul}(\vx), \mathcal M_{sl}(\vx))$ on $\sS$ using ERM, update $\mathcal M_{sl}, \mW_{sl}, h$ until converge.
        \ENSURE $\mathcal M_{FTT}$
	\end{algorithmic}  
\end{algorithm}

\textbf{Step 1. Unsupervised freeze stage.}
FTT starts with a model $\mathcal M_{init}$ pretrained in large datasets like ImageNet or language corpus.
Given a training set $\sS_{tr} \sim \sD_{tr}$, we use an unsupervised method like Contrastive Learning or Principal Component Analysis (PCA)
to learn $pm$ features, where $m$ is the number of total features, and $p\in[0,1]$ is a hyper-parameter denoting the fraction of unsupervised features. This stage gives a submodel $\mathcal M_{ul}: \mathcal X \mapsto \R^{pm}$, where $ul$ stands for ``unsupervised learning''.

\textbf{Step 2. Supervised train stage.}
Then, we freeze $\mathcal M_{ul}$ and train the other $(1-p)m$ dimensional features as well as a linear head together using a supervised method.
Specifically, we copy the initial pretrained checkpoint $\mathcal M_{init}$, i.e. we set $\mathcal M_{sl} = \mathcal M_{init}$. 
We set its output dimension to $(1-p)m$, and add a linear head $h$ upon $(\mathcal M_{ul}, \mathcal M_{sl})$ with input dimension $pm + (1-p)m = m$.
In this way, the complete network output is $\mathcal M_{FTT}(\vx) = h((\mathcal M_{ul}(\vx), \mathcal M_{sl}(\vx))$.
We supervisedly train $\mathcal M_{FTT}$ where only the parameters in $h$ and $\mathcal M_{sl}$ is optimized (with $\mathcal M_{ul}$ being frozen).

\subsection{Theoretical Guarantees of FTT}
We now show that in our two-layer network setting, FTT can guarantee a better probing performance than ERM under different non-realizable noises. 
Suppose in the freeze stage, the representation learned by PCA is $\tilde{\mW}_{ul}\in\R^{d\times pm}$. %Let us denote the features learned by PCA by $\tilde{\mW}_{PCA}$.
We similarly initialize $\mW_{sl}(t) \in \mathbb R^{d \times (1-p)m}, \vb(t)  = \begin{pmatrix}
    \vb_{ul}(t) \\ \vb_{sl}(t)
\end{pmatrix}\in \mathbb R^{m \times 1}$, and train $\mW_{FTT}(t) = (\mW_{ul}, \mW_{sl}(t))$ and $\vb(t)$.
Notice that $\mW_{ul}$ will not be updated.

\begin{theorem}[FTT Bound]\label{theorem_FTT}
    Suppose $p > \frac {k}{m}$. We still assume that throughout the whole optimization, $\|\mW_{sl}(t)\vb_{sl}(t)\|_2 > c_0 > 0 $. Then, for any time $t$, any $\eta_{core}^2 \not= -\eta_{spu}^2 \beta \left(\eta_{spu}^2 \mI - \mSigma\right)^{-1}\mSigma \beta$ (which is true a.s.),
    \begin{align}
        \ell_{te}(\mW_{FTT}(t)) \leq {\rm err}_{te}^* + \mathcal O(t^{-1}).
    \end{align}
\end{theorem}
\Cref{theorem_FTT} suggests that when we preserve enough unsupervised features, FTT can converge to the \textit{optimum} in $\ete$ for most of $\eta_{core}$ and $\eta_{spu}$. 
It can circumvent the lower bound in \Cref{theorem_lower} where one only uses ERM ($p = 0$);
it can also outperform the pure unsupervised method, since pure PCA features cannot attain $\text{err}_{tr}^*$ either.
It is by combining both features in $\mW_{ul} $ and $\mW_{sl}$ that FTT can surprisingly reach the optimum.
This effectiveness will be further verified by thorough experiments in the next section.

\subsection{Discussions on FTT}
\textbf{Selection of training algorithms.} 
Notice that FTT is a meta-algorithm, since it can be built on any supervised and unsupervised method.
To illustrate the effectiveness of our method, in this paper we simply use PCA in the ``freeze'' stage and ERM in the ``train'' stage.
This ensures that the effectiveness of FTT does not  take advantage of other algorithms that are carefully designed for these tasks.

\textbf{Selection of $p$.}
The unsupervised fraction $p$ is the only hyper-parameter. 
In terms of expressiveness, FTT is strictly stronger than a supervisedly trained model with features $(1-p)m$.
% , which is an initial guarantee of FTT.
We verify in \Cref{sec:pfraction} that FTT works well with various selection of $p$, e.g. between $[0.25,0.75]$.

\textbf{Computational cost.}
Although FTT is twice as large as the base model $\mathcal M_{init}$, in the supervised training stage the size of parameters to be optimized remains unchanged, since $\mathcal M_{ul}$ is frozen.
In practice, we find that the computation time and the GPU memory cost are indeed unchanged in each epoch.
For the ``freeze'' stage, we only conduct a PCA, which can be quickly done even in CPU.
For more discussions, please refer to Appendix~\ref{appendix:alg}.

%% file: Sections/Experiments.tex
\section{Experiments}
\label{sec:experiments}

In this section, we experimentally verify our theories in real-world datasets, compare FTT with other algorithms, and conduct ablations.\footnote{Our code can be found at \url{https://github.com/YWolfeee/Freeze-Then-Train}. %in \href{https://github.com/YWolfeee/Freeze-Then-Train}{this Github Repository}.
}
An overview of our experimental setup is provided below; see Appendix~\ref{appendix:details}  for more details.

\textbf{Noise generation.} 
To systematically study the influence of noise, we follow \cite{zhang2021learning} and explicitly generate noise by flipping labels. 
Notice that labels are noisy for all data we obtain, no matter what the training set $\sS_{tr} \sim \etr$ and the test-time probing set $\sS_{te} \sim \ete$ are.
Nevertheless, our goal is to recover the \textit{ground truth}.
To accurately evaluate the method, we further divide $\sS_{te}$ into a validation split $\sS_{val}$ and a testing split $\sS_{te}$. 
The labels are noisy in $\sS_{tr}$ and $\sS_{val}$, but are noiseless in $\sS_{te}$.
We retrain the last layer using only the validation split, and report performance on the testing split that is never seen.

\textbf{Datasets.}
We consider Waterbirds \citep{sagawa2019distributionally} and CelebA \citep{liu2015deep}, as well as Dominoes used in \cite{shah2020pitfalls,pagliardini2022agree}. 
% Some images are shown in [TODO].
\begin{itemize}
    \item
\textbf{Dominoes} is a synthesis dataset based on CIFAR10 and MNIST. The top half of the image shows CIFAR10 images (core features)
% from classes \{car, truck\} 
and the bottom half shows MNIST images (spurious features).
% from classes \{0,1\}.
% The core features are the CIFAR10 images while the spurious features are the MNIST digits.
Digits are spuriously correlated to labels in $\etr$, but are independent with labels in $\ete$.
Given a target core noise $\eta_{core}$ and spurious noise $\eta_{spu}$, we first randomly flip $\eta_{core}$ fraction of the ground truth in CIFAR to obtain $y_{core}$ and $\eta_{spu}$ fraction of the ground truth in MNIST to obtain $y_{spu}$.
For $\etr$, we concatenate CIFAR and MNIST images with the same label, i.e. $y_{core} = y_{spu}$.
For $\ete$, digits are randomly concatenated with CIFAR images.
% The final labels are $y_{core}$.
We select the $\eta_{core}$ and $\eta_{spu}$ separately from $\{0,5,10,15,20\}$ ($\%$), resulting in 25 settings of noise.
    \item 
    \textbf{Waterbirds} is a typical spurious correlation benchmark. 
    The label is the type of bird (water-bird $= 0$ or ground-bird $=1$), which is spuriously correlated with the background (water $=0$ or ground $=1$).
    In the training split $\sS_{tr}$, the spurious noise $\eta_{spu}$ is $5\%$, while in $\sS_{val}$ and $\sS_{te}$ we have $\eta_{spu} = 50\%$. 
    % In Waterbirds we cannot tune the feature noise freely.
    % To this end, we keep the spurious noise unchanged, and tune the core noise by flipping labels.
    Given $\eta_{core}$, we flip the label of the dataset according to \Cref{tab:flip}. For example, we select $\textcolor[rgb]{0, 0, 0.545}{\frac p 2 \eta_{core}}$ fraction of data from $(0,0)$ and flip labels to $1$. This will increase $\eta_{core}$ and $\eta_{spu}$ by $\frac p 2 \eta_{core}$.
    We also select $\textcolor[rgb]{0.545, 0, 0}{\frac p 2 \eta_{core}}$ fraction of data from $(0,1)$ and flip labels to $1$. This will increase $\eta_{core}$ but decrease $\eta_{spu}$ by $\frac p 2 \eta_{core}$. 
    Similarly, we flip $\frac {1-p} 2 \eta_{core}$ fraction of data with label $1$.
    After flipping, the spurious noise is kept unchanged, but the core noise increases from $0$ to $\eta_{core}$. 
    We select $\eta_{core}$ from $\{0, 2,4,6,8,10\}$ in percentage. The spurious noise in the $\sS_{tr}, \sS_{val}, \sS_{te}$ is $5\%, 50\%, 50\%$.
    \begin{table}[h]
    \small
        \centering
        \begin{tabular}{c|cc}
          \makecell{(Core, Spurious)}  &  \makecell{Origin\\fraction} & Flip fraction\\
        \hline
        $(0, 0)$ & $p_0s$ & $\textcolor[rgb]{0, 0, 0.545}{-\frac {p_0} 2 \eta_{core}}  \textcolor[rgb]{0.545, 0, 0}{+\frac {p_1} 2 \eta_{core}}$  \\
        $(0, 1)$ & $p_0(1-s)$ &  $\textcolor[rgb]{0.545, 0, 0}{-\frac {p_0} 2 \eta_{core}} \textcolor[rgb]{0, 0, 0.545}{+\frac {p_1} 2 \eta_{core}}$\\
        $(1, 0)$ & $p_1(1-s)$ &$\textcolor[rgb]{0, 0, 0.545}{+\frac {p_0} 2 \eta_{core}} \textcolor[rgb]{0.545, 0, 0}{-\frac {p_1} 2 \eta_{core}}$\\
        $(1, 1)$ & $p_1s$& $\textcolor[rgb]{0.545, 0, 0}{+\frac {p_0} 2 \eta_{core}} \textcolor[rgb]{0, 0, 0.545}{-\frac {p_1} 2 \eta_{core}}$\\
        % ground-bird & (1-p)s & (1-p)(1-s)\\
        % \hline
        \end{tabular}
        \caption{The fraction of data to be flipped to generate core noise $\eta_{core}$ in Waterbirds and CelebA.
        The (Core, Spurious) column represents the label of the core feature and the spurious feature.
        $p_0, p_1$ is the fraction of data with label $0,1$, and $s$ is the spurious correlation.}
        \vspace{-10pt}
\label{tab:flip}
    \end{table}
        \item
    \textbf{CelebA} is a binary classification dataset, where the label is the color of hair (non-blond $=0$ or blond $=1$), and is spuriously correlated with the gender (female $=0$ or male $=1$).
    The major difference between CelebA and Waterbirds is that the spurious noise in CelebA is large ($42\%$).
    To better study the probing performance under different noises, we drop a fraction of data with (color, gender) $=(0,0)$ such that $\eta_{spu}$ in $\sS_{tr}$ is kept to $6\%$ within data groups with label $0$ and $1$.
    The label flipping process is the same as in Waterbirds.
    % , and $\eta_{core}$ is selected from $\{0, 2,4,6,8,10\}$.
    % , and flip the labels according to \Cref{tab:flip}.
    % The spurious correlation
    
% In Waterbirds and CelebA, the noise of spurious features are already fixed ($5\%$ and $42\%$). To this end, we keep the 
\end{itemize}

\begin{table*}[th]
\begin{center}
\small
\begin{tabular}{ccccccccccccc}
\hline
\multirow{2}{*}{Dataset} & \multirow{2}{*}{$\eta_{core}$ (\%)} & \multicolumn{5}{c}{Worst Group Accuracy (\%)} & & \multicolumn{5}{c}{Average Accuracy (\%)} \\
\cline{3-7} \cline{9-13}
&&ERM&IRM&CVaR-DRO &JTT&Ours & &ERM&IRM&CVaR-DRO&JTT& Ours\\
\hline
\multirow{7}{*}{Waterbirds}
& 0 & 95.0 & \textbf{95.3} & 94.3 & 93.3 & 94.5 &
& 95.3 & \textbf{95.5} & 94.6 & 94.1 & 94.9 \\

& 2 & 93.6 & \textbf{94.1} & 93.8 & 89.7 & 93.6 &
& 94.2 & \textbf{94.3} & 94.0 & 90.7 & 94.2 \\

& 4 & 92.8 & 92.8 & 92.8 & 85.3 & \textbf{92.9} &
& 93.2 & \textbf{93.5} & 93.2 & 85.9 & \textbf{93.5} \\

& 6 & 90.8 & 91.5 & 77.8 & 86.8 & \textbf{92.8} &
& 91.3 & 91.8 & 77.8 & 87.1 & \textbf{92.9} \\

& 8 & 88.5 & 88.8 & 77.8 & 82.0 & \textbf{92.7 }&
& 89.9 & 90.1 & 77.8 & 82.7 & \textbf{93.0} \\

& 10 & 87.6 & 87.9 & 77.8 & 78.6 & \textbf{92.4} &
& 89.4 & 89.4 & 77.8 & 78.9 & \textbf{92.9} \\
\cline{2-13}
& Mean & 91.4 & 91.7 & 85.7 & 86.0 & \textbf{93.1} &
& 92.2 & 92.4 & 85.9 & 86.6 & \textbf{93.6} \\

\hline
\multirow{7}{*}{CelebA} 
& 0 & 95.0 & 95.2 & 92.9 & 94.4 & \textbf{95.3} &
& \textbf{97.2} & \textbf{97.2} & 96.0 & 96.7 & \textbf{97.2} \\

& 2 & \textbf{95.2} & \textbf{95.2} & 92.4 & 91.6 & \textbf{95.2} &
& \textbf{97.2} & \textbf{97.2} & 95.9 & 96.0 & \textbf{97.2} \\

& 4 & 94.5 & 94.2 & 91.9 & 92.7 & \textbf{94.9} &
& 97.1 & 97.0 & 95.5 & 96.4 & \textbf{97.2} \\

& 6 & 94.3 & 94.3 & 91.5 & 92.0 & \textbf{94.4} &
& 96.9 & 96.9 & 95.5 & 96.0 & \textbf{97.0} \\

& 8 & 93.7 & 93.8 & 91.4 & 91.4 & \textbf{94.0} &
& \textbf{96.7} & \textbf{96.7} & 95.4 & 95.7 & \textbf{96.7} \\

& 10 & 92.4 & 92.8 & 91.1 & 80.5 & \textbf{93.1} &
& 96.2 & 96.2 & 95.4 & 92.1 & \textbf{96.3} \\

\cline{2-13}
& Mean & 94.2 & 94.2 & 91.9 & 90.4 & \textbf{94.5} &
& \textbf{96.9} & \textbf{96.9} & 95.6 & 95.5 & \textbf{96.9} \\
\hline
\end{tabular}
\caption{
Test-time probing accuracy (\%) for four methods on Waterbirds and CelebA, under different core noises $\eta_{core}$.
\textbf{Bold} means the best accuracy across four methods.
The ``Mean'' row stands for the average accuracy across $\eta_{core}$.
We repeat all settings $10$ times and average the numbers.
For worst group accuracy, FTT (ours) can be competitive when $\eta_{core}$ is small and outperform other algorithms by at most $4.5\%$ when $\eta_{core}$ increases.
It can increase accuracy by $1.4\%$ and $0.3\%$ on Waterbirds and CelebA on average.
}
\end{center}
\label{tab:main}
\vspace{-10pt}
\end{table*}

\textbf{Models.} 
For Dominoes we use ResNet18 \citep{he2016deep}, and for Waterbirds and CelebA we use ResNet50.
We load ImageNet pretrained weights \citep{tanaka2018joint} from \verb|torchvision.models| \citep{paszke2017automatic}.
% We specify other training settings in the appendix.
% We train the model with \verb|torch.optim.SGD| with \verb|momentum_decay| set to $0.9$. 
% The N\_epochs, the weight decay value and the learning rate are datasets specific, and will be specified in the appendix.

\textbf{Methods.}
%\textcolor{red}{Updated. }
We compare FTT with ERM, IRM \citep{arjovsky2019invariant}, CVaR-DRO \citep{duchi2019distributionally} and JTT \citep{jtt}. 
IRM is a widely used OOD generalization algorithm, and CVaR-DRO and JTT are competitive robust training methods that perform well in several benchmark datasets for studying spurious correlations. %when core features are perfect.
% However, they are not tested under both spurious correlations and feature noise.
For IRM, we use hyperparameters in  We use hyperparameters in \cite{domainbed}. For CVaR-DRO and JTT, we use hyperparameters searched in \cite{jtt}.
For ERM and test-time probing, we use parameters in \cite{lastlayer}.
For FTT, we set $p = 0.25$.
% except in \Cref{sec:pfraction} where we conduct ablation experiments.
% on $p$ in \Cref{sec:pfraction}.
% Notice that for most mild selection of $p$ (e.g. in $[0.25, 0.75]$) FTT works quite well.
% Notice that for all methods, the group information is never used during training in $\etr$.

\textbf{Test-time Probing.}
% Now we formally specify the whole experiments pipeline.
% For each method and each dataset, we first train the model using this method in the training split of the dataset.
After we train a model in $\sS_{tr}$, we need to retrain the last layer in $\sS_{val}$.
We follow \cite{lastlayer} and divide $\sS_{val}$ into two subsets, where the first subset is used to retrain the last layer, and the second is to select hyperparameters.
Specifically, We sub-sample the first subset using the group information such that the data population from each of the two groups are identical\footnote{Previous works divide binary datasets into 4 groups according to both labels and spurious features. However, under non-realizable noises, manually splitting groups according to possibly incorrect labels become meaningless. We only consider two groups defined across spurious features. This setting is kept for all experiments and methods to make sure the comparison is fair.}.
We then perform logistic regression on this sub-sampled dataset. 
% Notice that we consider 2 groups rather than 4.
This process is repeated for 10 times, and we average these learned linear weights and obtain the final last layer weight and bias.
We then use the second group to select the hyperparameters, i.e. the regularization term $C$ according to the worst spurious group accuracy.
After probing, we save the model and evaluate the worst group accuracy and the average accuracy in the test split where the label is noiseless.
For Dominoes, each setting is repeated for 5 times; for Waterbirds and CelebA 10 times.
Each reported number is averaged across these runs.

\begin{figure}[t]
\begin{center}
\includegraphics[width=0.45\textwidth]{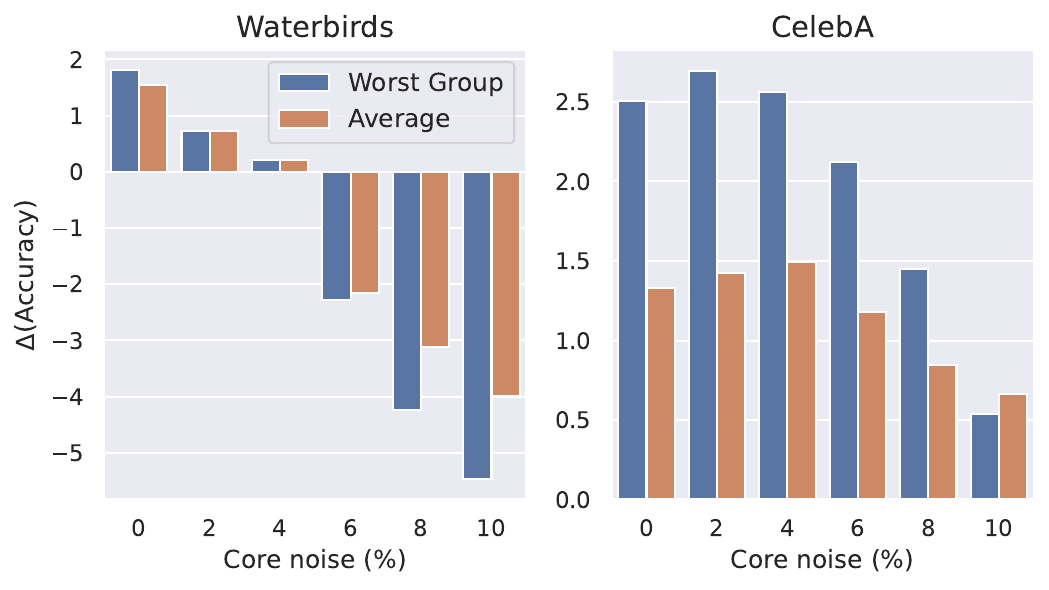}
\end{center}
\vspace{-10pt}
\caption{
Test-time probing accuracy gap between trained model $\mathcal M_{erm}$ and initialized model $\mathcal M_{init}$ on Waterbirds and CelebA.
The x-axis is the core noise and the y-axis is the improvement of accuracy.
In both datasets, the improvement of both worst group accuracy and average accuracy decrease when $\eta_{core}$ increases.
In Waterbirds, large $\eta_{core}$ can even make ERM training harmful.
}
\label{fig:noise_matter}
\vspace{-10pt}
\end{figure}

\subsection{Examine Non-realizable Noise Theories}
% We now examine our theories in three datasets.

\paragraph{Noise matters in Dominoes.} 
% We train a ResNet-18 model using standard ERM for 200 epochs, and save the trained model $\mathcal M_{erm}$ as well as the initialized model $\mathcal M_{init}$.
% Under different $\eta{core}$ and $\eta_{spu}$, we show the gap of probing accuracy for $\mathcal M_{erm}$ and $\mathcal M_{init}$ in \Cref{fig:dominoes_erm}.
We compare the test-time probing accuracy gap between the ERM-trained model $\mathcal M_{erm}$ and initialized model $\mathcal M_{init}$ under different noises in \Cref{fig:dominoes_erm}.
When $\eta_{core}<\eta_{spu}$ (the upper-triangle part), the probing accuracy improved by $6.7\%$ (both for the worst group and in average).
However, as $\eta_{core}$ increases or $\eta_{spu}$ decreases, this accuracy improvement diminishes from $6.7\%$ to $-12\%$.
The trends are clear if we consider any certain row or column, where $\eta_{core}$ ($\eta_{spu}$) is fixed and $\eta_{spu}$ ($\eta_{core}$) alters.
Despite ERM learns core features when $\eta_{core} < \eta_{spu}$, it \textit{cannot} preserve them when $\eta_{core} > \eta_{spu}$.
% This matches our theories and prove the 

\paragraph{Noise matters in Waterbirds and CelebA.} 
We now turn to Waterbirds and CelebA.
% to further examine our theories.
% We train a ResNet-50 mdodel using ERM for 100 epochs in Waterbirds and 50 epochs in CelebA. 
We similarly save $\mathcal M_{erm}$ as well as $\mathcal M_{init}$, calculate their probing accuracies, and show the gap in \Cref{fig:noise_matter}.
In both datasets, test-time probing accuracy decreases when $\eta_{core}$ increases.
For instance, for worst group accuracy, the improvement is $1.81\%$ for Waterbirds and $2.5\%$ for CelebA when $\eta_{core} = 0$, but becomes $-5.5\%$ for Waterbirds and $0.5\%$ for CelebA when $\eta_{core} = 10\%$.
% This is in comparison with the initialized model, and 
% This implies that ERM cannot improve the quality of features and can even do harm to them when core feature noise is relatively larger. 
In Waterbirds, ERM becomes detrimental even when $\eta_{core} = 6\%$ is slightly larger than  $\eta_{spu} = 5\%$.
% This again supports our theories, and call for a better method to learn better features both when $\eta_{core}$ is small or large compared with $\eta_{spu}$.

% \subsection{FTT Outperforms Other Methods}
\subsection{Effectiveness of FTT}
\label{sec:main_table}

\subsubsection{Spurious Correlation Benchmarks}
We now compare FTT with other algorithms,
% We run JTT and CVaR-DRO in Waterbirds and CelebA with different $\eta_{core}$, using the benchmarks in \cite{jtt}. 
% We save the best model according to their model selection strategy.
% For ERM we run the standard benchmark in \cite{lastlayer}.
% All methods run 100 epochs in Waterbirds and 50 epochs in CelebA.
% After saving models, we use the same DFR benchmark to retrain the last layer to ensure the comparison is fair.
and show results in \Cref{tab:main}. 
For the worst group accuracy, FTT attains $93.1\%$ in Waterbirds and $94.5\%$ in CelebA on average, outperforming ERM and other robust training algorithms by $1.4\%$ and $0.3\%$. 
When $\eta_{core}$ is small, purely supervised methods can perform quite well, and FTT can match their performance.
When $\eta_{core}$ increases, purely supervised based algorithms are biased to learn more spurious features, 
% which results in a decrease in probing accuracy. 
% However, since FTT preserves salient features during unsupervised training, 
while FTT can resist non-realizable noises during training.
% , and use the preserved features to perform well in $\ete$.
In waterbirds, it can recover accuracy by $4.5\%$.

An interesting observation is the performance of CVaR-DRO and JTT. 
They are robust training algorithms that intuitively emphasize the importance of samples that are incorrectly classified. 
It turns out that this focus could be misleading where there exist non-realizable noises, since the emphasized samples can be classified wrong because of the noise.
% As a result, weighting more on those samples mislead the model.
In Waterbirds, they perform nearly $10\%$ worse than ERM, suggesting that relying too much on labels might backfire in situations where we do not know if features can be noisy.
On the contrary, FTT overcomes this problem by finding features in an unsupervised way.

We also compare FTT with ERM in Dominoes, and show results in Appendix~\ref{appendix:more_results}. Averaged across 25 noise settings, FTT attains $78.1\%$ (worst group) and $78.8\%$ (average), outperforming ERM by $4.1\%$ and $4.0\%$.
Together, FTT shows the ability under different noises, overcoming the drawback of ERM when $\eta_{core}$ is large.

\subsubsection{General Distribution Shift Benchmarks}

To further illustrate the effectiveness of FTT, we consider more general distribution shift benchmarks, where there is no explicit spurious correlation and explicit noise between features and labels. Specifically, we consider three OOD multi-class classification datasets: PACS with 7 classes \citep{li2017deeper}, Office-Home with 65 classes \citep{venkateswara2017deep}, and VLCS with 5 classes \citep{torralba2011unbiased}.
Each dataset has four domains, and images in different domains have different styles, e.g. sketching, painting, or photography.
The task is to train a model on three domains, and perform well in the unseen test domain.
Following the last layer retraining setting, we also allow the model to retrain the last linear layer on the unseen test domain, i.e. we still consider the retraining accuracy.

 \begin{table}[tb]
 \small

    \begin{center}
        
    \begin{tabular}{ccccccc}
         \hline
        \multirow{5}*{PACS} & Domain & A & C & P & S&Mean\\
        \cline{2-7}
        & ERM & 89.2 & 93.2 & 95.8 & 88.4 & 91.7 \\
        & IRM & 61.1 & 67.5 & 81.7 & 79.1 & 72.4 \\
        & DRO & 91.9 & 92.7 & 95.8 & \textbf{91.3} & 93.0 \\
        & Ours & \textbf{92.7} & \textbf{94.9} & \textbf{97.9} & 90.8 & \textbf{94.1} \\
        
        \hline
        \multirow{5}*{\makecell[c]{Office-\\Home}} & Domain & A & C & P & R&Mean\\
        \cline{2-7}
        & ERM & 69.9 & 69.9 & 87.8 & 78.8 & 76.6 \\
        & IRM & 25.2 & 44.9 & 69.3 & 54.0 & 48.3 \\
        & DRO & 72.8 & 73.1 & \textbf{88.5} & 79.9 & 78.6 \\
        & Ours & \textbf{73.8} & \textbf{73.7} & 87.1 & \textbf{83.1} & \textbf{79.4} \\
        \hline
        
        \multirow{5}*{VLCS} & Domain & C & L & S & V&Mean\\
        \cline{2-7}
        & ERM & 99.3 & 75.0 & 77.3 & 81.5 & 83.3 \\
        & IRM & 75.3 & 62.5 & 59.5 & 60.4 & 64.4 \\
        & DRO & 99.6 & 74.0 & 78.5 & 81.8 & 83.5 \\
        & Ours & \textbf{100.0} & \textbf{76.6} & \textbf{81.1} & \textbf{84.6} & \textbf{85.6} \\
       
        \hline
        
    \end{tabular}
    \caption{Test-time probing accuracy (\%) for 4 methods on PACS, Office-Home, and VLCS. Rows ``Domain'' specify which domain among 4 domains is unseen during the training stage, therefore used to retrain the last layer. The ``Mean'' column stands for the average accuracy across different test domain selections, and we \textbf{bold} the highest accuracy among 4 methods in each setting.
    FTT (ours) consistently outperforms other methods by 1.1\% on PACS, 0.8\% on Office-Home, and 2.1\% on VLCS.}
    \label{tab:distshift}
    \end{center}
    \vspace{-10pt}
\end{table}

We compare FTT with ERM, IRM, as well as GroupDRO\citep{sagawa2019distributionally}, and we use the implementation and hyperparameters in \cite{domainbed}. 
Specifically, for each dataset and each domain as the test domain, we use the default settings (for FTT, $p = 0.25$) to train a model using each algorithm on the rest three domains, retrain the last layer on the test domain using linear regression, and report the accuracy.
Notice that GroupDRO is different from CVaR-DRO where the latter does not rely on group information. 
We report all numbers in Table~\ref{tab:distshift}.

Across three datasets, 12 test domain settings, FTT consistently outperforms all other methods by 1.3\% on average.
Importantly, FTT is initially designed to remove spurious correlations, which is a special type of OOD generalization. However, we find that it also works well in general OOD settings such as in distribution shift datasets, showing that FTT is robust and effective.

\vspace{-0.3cm}
\subsection{Ablation Studies}
% In this section, we conduct ablations on FTT.
\vspace{-0.2cm}
\subsubsection{Selection of p (unsupervised fraction)}
\label{sec:pfraction}
\begin{table}[!h]
\small
\begin{center}
\begin{tabular}{cccccc}
% \hline
\multirow{2}{*}{\makecell{$\eta_{core}$ (\%)}}  & \multicolumn{5}{c}{unsupervised features fraction ($p$)} \\
\cline{2-6}
&0.00&0.25&0.5&0.75&1.00 \\
\hline
0 & \textbf{95.0} & 94.5 & 94.8 & 94.6 & 93.2 \\
2 & 93.6 & 93.6 & \textbf{94.3} & 94.1 & 92.9 \\
4 & 92.8 & 92.9 & 93.2 & \textbf{93.7} & 92.6 \\
6 & 90.8 & 92.8 & \textbf{93.3} & \textbf{93.3} & 93.0 \\
8 & 88.5 & 92.7 & 92.6 & \textbf{93.0} & 92.7 \\
10 & 87.6 & 92.4 & 93.1 & 93.0 & \textbf{93.1} \\
\hline
Mean & 91.4 & 93.1 & \textbf{93.6} & \textbf{93.6} & 92.9 \\

\end{tabular}
\end{center}
\vspace{-5pt}
\caption{
Worst group accuracy on Waterbirds, under different $p$. The setting is the same as \cref{tab:main}.
For all $p \in [0.25,0.75]$, FTT outperforms ERM by at least $1.7\%$. 
We find that the best $p$ value increases as $\eta_{core}$ increases.}
\label{tab:p_value}
\vspace{-5pt}
\end{table}
FTT is a simple but effective framework, 
% and it can be built on any existing algorithm, 
where the only hyperparameter is the fraction of unsupervised features $p$.
We now compare the worst group accuracy of FTT on Waterbirds under different $p$ values in \Cref{tab:p_value}.
When $p=0$, FTT is the same as ERM;
when $p=1$, FTT is the same as PCA.
% We show the worst group accuracy in Waterbirds in \Cref{tab:p_value}.
We find that FTT is relatively insensitive to $p$, with that no matter $p = 0.25, 0.5, 0.75$, FTT can consistently outperform ERM by at least $1.7\%$.
On the other hand, we do find that as the noise increases, a more ``unsupervised'' method is favored, which matches our expectation.
The ablation on other datasets can be found in Appendix~\ref{appendix:more_results}.

\subsubsection{Number of features}
% \vspace{-0.3cm}
\begin{figure}[h]
\begin{center}
\includegraphics[width=0.45\textwidth]{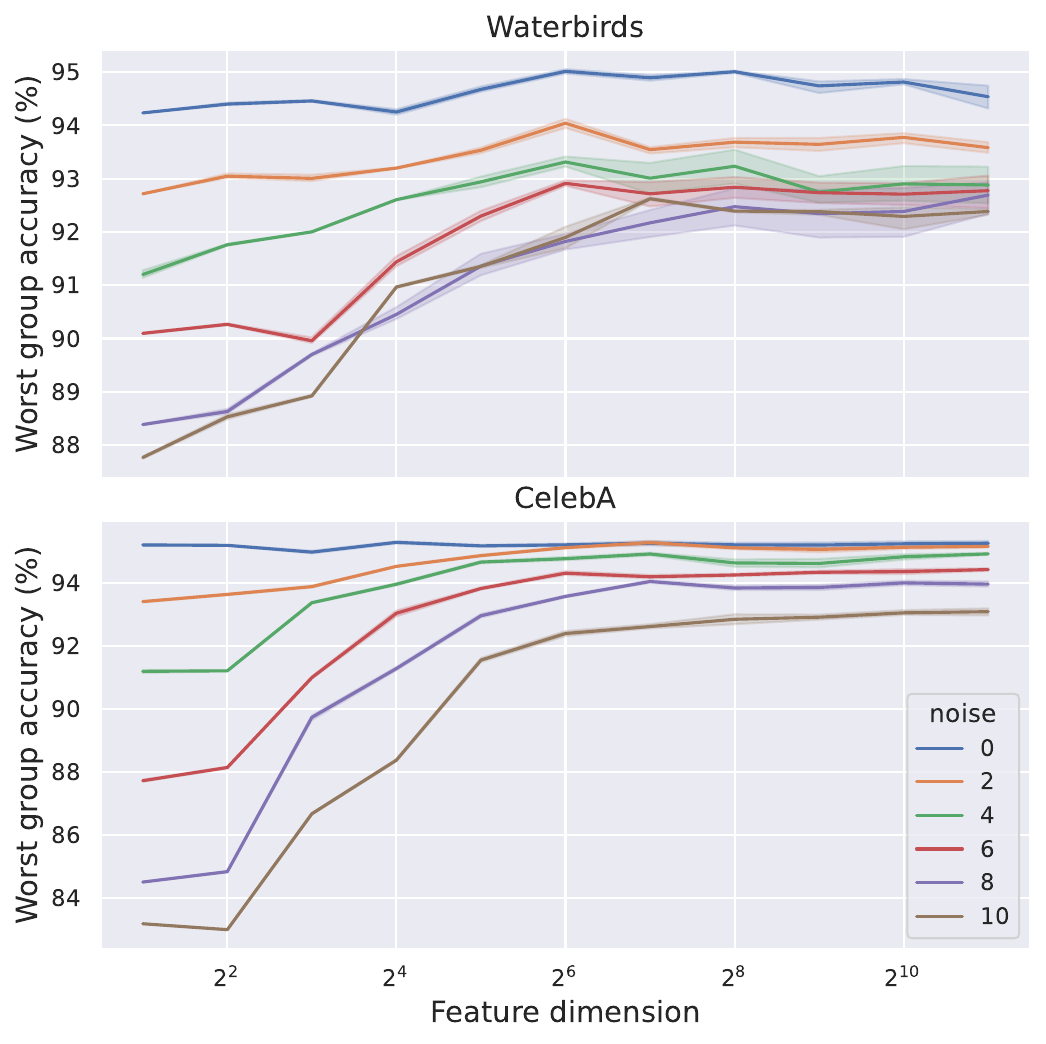}
\end{center}
\vspace{-10pt}
\caption{
Worst group accuracy on Waterbirds and CelebA for FTT. 
The x-axis is the feature dimension in log scale.
}
\label{fig:dimension_waterbirds}
\vspace{-5pt}
\end{figure}

How many features do we actually need to make last layer retraining work?
This is important since in $\ete$ the computation resource is limited, and preserving too many features is impractical.
To this end, we use PCA to project the features that are learned in $\etr$, and then retrain the last layer on the low-dimensional features.
Since PCA does not require group information (not even labels), it can be accomplished in $\etr$.
We consider the projection dimension varying from $2^1$ to $2^{12}$, and show results in \Cref{fig:dimension_waterbirds}.
After training on $\etr$, only a few features are enough to perform well (or even better) in $\ete$. 
Averaged across different noise settings, FTT attains $93.2\%$ on Waterbirds and $94.2\%$ on CelebA when $m = 64$, matching $93.1\%$ and $94.5\%$ when using all features, and speeding up the probing process $3.9$ times.
This suggests that FTT is computational friendly in test-time probing, and the improvement is significant.

%% file: Sections/Conclusion.tex
\vspace{-0.2cm}
\section{Conclusions}
\vspace{-0.2cm}
In this paper, we study the test-time probing strategy as a way to overcome spurious correlations. 
We theoretically and empirically show that ERM recovers core features only when the non-realizable noise of core features is much smaller than the that of spurious features.
We propose FTT to overcome this problem and outperform other algorithms under different settings.
Our work suggests that by properly combining unsupervised and supervised methods, machine learning models can be more robust and accurate to spurious correlations.

%% file: Sections/supplement.tex
\section{More Discussion on FTT}
\label{appendix:alg}
In practice, for the unsupervised learning, we conduct PCA using \verb|sklearn.decomposition.PCA|. 
When the population of $\sS_{tr}$ is too large, we can randomly sub-sample the dataset before PCA, which does not influence the quality of the learned features.
Notice that when the initialized model $\mathcal M$ is random (e.g. rather than an ImageNet pretrained model), pure PCA will not work.
In this case, we can consider any unsupervised training method, like the Contrastive Learning algorithm \citep{zhang2022correct}.
However, if these unsupervised methods still fail to extract core features (such as when the core features are too complex for any unsupervised methods to learn), FTT might degrade to simple ERM.

After the unsupervised training, we reinitialize a model $\mathcal M_{sl}$ with $(1-p)m$ features.
During the supervised training, we concatenate $pm$ unsupervised features and $(1-p)m$ supervised features, apply a linear layer on the $m$ features to obtain $K$ outputs where $K$ is the number of classes, and compute the cross entropy loss with labels. We only update parameters in the last linear layer and the supervised model (with $(1-p)m$ features), while the unsupervised model $\mathcal M_{ul}$ is kept unchanged.
As a result, the training time remains unchanged during the supervised training, since the number of parameters to be optimized remains unchanged.

\section{Experimental Details}
\label{appendix:details}
In this section, we give details on how we implement our experiments.

\subsection{Benchmarks}
As mentioned in the main paper, we consider Dominoes, Waterbirds and CelebA, which is the same as in \cite{lastlayer}.
For each dataset, the core feature and the spurious feature are different.
In Dominoes, the core feature is the CIFAR image (\verb|car = 0, truck = 1|), and the spurious feature is the MNIST digits (\verb|zero = 0, one = 1|).
In Waterbirds, the core feature is the type of bird (\verb|water-bird = 0, ground-bird = 1|), and the spurious feature is the background (\verb|water = 0, ground = 1|).
In CelebA, the core feature is the color of the hair of the person in the image (\verb|Non-blond = 0, Blond = 1|), while the spurious feature is the gender of the person (\verb|Female = 0, Male = 1|).

The number of data we use for each split is shown in \Cref{tab:data_num}. Notice that this table shows the dataset when no noise is explicitly added.
In Dominoes, the spurious correlation is perfect in $\sS_{tr}$ but is complete broken in $\sS_{val}$ and $\sS_{te}$. 
    Tn Waterbirds the spurious correlation in $\sS_{tr}$ is 95\% ($\eta_{spu} = 5\%$), while in $\sS_{val},\sS_{te}$ it is random.
    In CelebA the situation is different. The spurious correlation is almost maintained in $\sS_{val}$ and $\sS_{te}$ (only slightly different in decimal point). This suggests that in terms of average accuracy in $\sS_{te}$, pure ERM should be able to work quite well, which is verified in \cref{tab:main}. Notice that the original population for $(0,0)$ in $\sS_{tr}$ is $71629$, and we drop most of them to create a small spurious noise for our study.
    Specifically, we calculate the spurious correlation within data with label $1$, which is $94.3\%$. We then select data from group $(0,0)$ sequentially until we get $x$ data such that $x / (x+66874) = 94.3\%$.

\begin{table}[h]
\small
    \centering
    \begin{tabular}{cccccccccc}
    \multirow{2}{*}{\makecell{(Core,\\Spurious)}} & 
    \multicolumn{3}{c}{Dominoes} &
    \multicolumn{3}{c}{Waterbirds} &
    \multicolumn{3}{c}{CelebA} \\
    \cmidrule(r){2-4} \cmidrule(r){5-7} \cmidrule(r){8-10} 
    & $\sS_{tr}$ & $\sS_{val}$ & $\sS_{te}$ 
    & $\sS_{tr}$ & $\sS_{val}$ & $\sS_{te}$ 
    & $\sS_{tr}$ & $\sS_{val}$ & $\sS_{te}$ \\
    \hline
    (0, 0) 
    & 5000 (50) & 2500 (25) & 500 (25) 
    & 3498 (73) & 467 (39)  & 2255 (39)
    & 4053 (4)  & 524 (4)   & 546 (2) \\
    
    (0, 1) 
    & 0 (0)     & 2500 (25) & 500 (25) 
    & 184 (4)   & 466 (39)  & 2255 (39)
    & 66874 (70)& 8276 (70) & 7535 (2) \\
    
    (1, 0) 
    & 0 (0)     & 2500 (25) & 500 (25) 
    & 56 (1)    & 133 (11)  & 642 (11)
    & 22880 (24)& 2874 (24) & 2480 (23)\\
    
    (1, 1) 
    & 5000 (50) & 2500 (25) & 500 (25) 
    & 1057 (22) & 133 (11)  & 642 (11)
    & 1387 (2)  & 182 (2)   & 180 (2) \\

    \hline
    
\end{tabular}
    \caption{The number of data for each (core feature, spurious feature) group in Dominoes, Waterbirds and CelebA. Each cell shows the population and (the proportion in percentage). }
    \label{tab:data_num}
\end{table}

\subsection{Noise Generation}
We now explain how we generate feature noise in detail.
Dominoes is a synthesis dataset where we can manipulate the label and concatenate features.
While in Waterbirds and CelebA this is impossible.
Therefore, their noise generation mechanism is different.

\paragraph{Dominoes noise generation.}
Assume we are given the original CIFAR dataset $\sS_{CIFAR}$ and MNIST dataset $\sS_{MNIST}$, and we want to generate a spurious correlation Dominoes dataset with core noise $\ecore$ and $\espu$.
To this end, we first randomly flip $\ecore$ fraction of labels in $\sS_{CIFAR}$ and $\espu$ fraction of labels in $\sS_{MNIST}$.
Then, we randomly concatenate CIFAR and MNIST images so long as their (possibly incorrect labels) are the same. 

\paragraph{Waterbirds and CelebA noise generation.}
In these two real-world datasets, we cannot randomly concatenate features. To this end, we only tune the core noise and keep the spurious noise unchanged.
When a sample from $(0,0)$ is flipped to $(1,0)$, the noise of both the core feature and spurious feature increases; on the other hand, when a sample from $(0,1)$ is flipped to $(1,1)$, the core noise increases while the spurious noise decreases.
We leverage this property to maintain $\espu$ while tuning $\ecore$, as shown in \cref{tab:flip}.

\subsection{Optimization}
Our experiments consist of two stages, train a model, and retrain the last layer (test-time probing). In this section, we specify the parameters in the training stage.

\paragraph{Dominoes.}
In Dominoes we only compare ERM and FTT.
We start with the pretrained ResNet-18 model and follow the training settings in \cite{shah2020pitfalls}.
We use SGD with \verb|weight_decay = 1e-3| and \verb|lr = 0.01| and train the model for $200$ epoch. We reduce \verb|lr| to $0.002$ after 50 epoch implemented by \verb|optim.lr_scheduler.LambdaLR|. The batch size is set to $256$.
We use the cross entropy loss implemented by \verb|F.cross_entropy|.

\paragraph{Waterbirds and CelebA.}
For these two datasets, we follow the implementation in \cite{lastlayer} for algorithm ERM and FTT, and follow the implementation in \cite{jtt} for JTT and CVaR DRO in order to make sure the model is trained well.
For ERM and FTT, we use SGD with \verb|momentum_decay = 0.9, lr = 1e-3| to train ResNet-50 models.
For waterbirds we use \verb|weight_decay = 1e-3|, and for CelebA we use \verb|weight_decay = 1e-4|. We train the model for 100 epochs in Waterbirds and 50 epochs in CelebA, and the batch size is set to $128$.
For JTT and CVaR DRO, we use the hyperparameters in \cite{jtt}.
Both methods use \verb|momentum_decay = 0.9, weight_decay = 1.0| on Waterbirds and \verb|momentum_decay = 0.9, weight_decay = 0.1| on CelebA.
For CVaR DRO on Waterbirds, the learning rate is set to \verb|1e-4|, and the alpha rate is set to \verb|0.2|; on CelebA the learning rate is \verb|1e-5| and the alpha rate is $0.00852$.
For JTT, according to their paper, an ERM model is trained for $T$ epochs first, and some data samples are up-weighted. Then, another ERM model will be trained using these data. All hyperparameters are inherited from their paper.
For these two algorithms, we use the best model according to their model selection method, i.e. the accuracy of a preserved validation set.

Once the training is finished, we will obtain a learned model where the final linear layer has input dimension $m$ and output dimension $2$. This layer will be removed and the Logistic Regression will be conducted on the $m$ dimensional features, as specified below.

\subsection{Test-time Probing}
We follow the deep feature reweighting algorithm to retrain the last layer.
Specifically, assume we are given $\sS_{val}$ that is sampled from $\ete$ (there is still noise).
We will use this dataset to retrain the last layer.
Specifically, we first down-sample a balanced dataset, i.e. the population of groups with different spurious labels are the same. We then split this down-sampled dataset into two parts. We train the last layer using the first part, and evaluate the performance on the second part.
Using the evaluated accuracy, we select the hyperparameter, i.e. the inverse regularization term $C$ in \verb|LogisticRegression|.
Finally, we fix the value of $C$, randomly sample $10$ balanced sets from $\sS_{val}$, train the weight and the bias for each set, and average across them.
This will be our final last layer.
We do NOT use \verb|solver = liblinear| and \verb|penalty = 'l1'|, since we empirically found that this cannot improve the performance much, but will slow down the retraining a lot.

Once the probing is done, we evaluate the performance on $\sS_{te}$, where the label is noiseless such that the numbers can accurately reflect the performance.

\begin{figure}[!h]
\begin{center}
\includegraphics[width=0.5\textwidth]{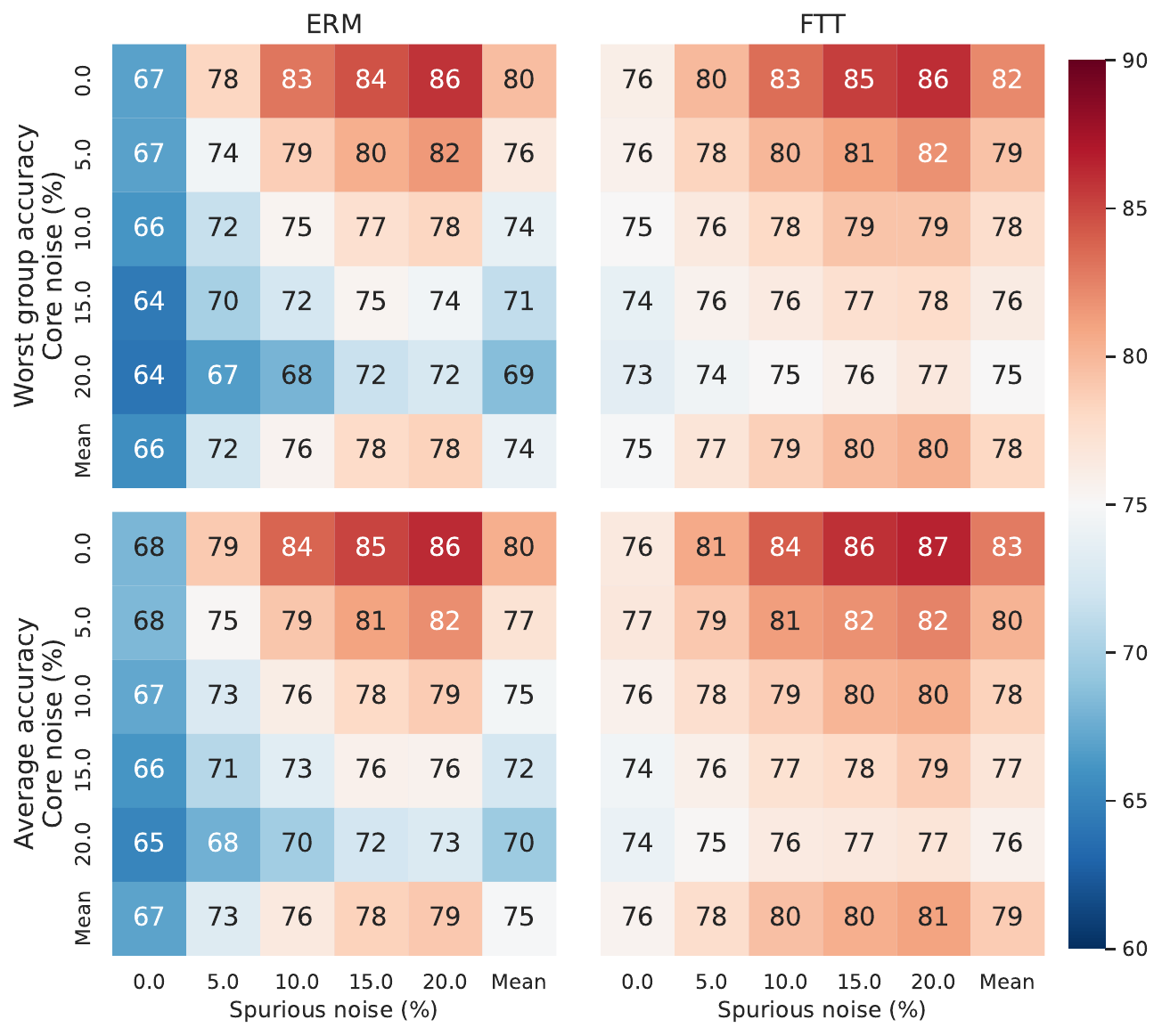}
\end{center}
\caption{
ERM (left) and FTT (right) worst group accuracy (top) and average accuracy (down) on Dominoes dataset, under different selection of $\eta_{core}$ and $\eta_{spu}$.
The ``Mean'' row and column stand the average across core noise and spurious noise separately.
}
\label{fig:dominoes_erm+ftt}
% \label{fig:noise_matter}
% \vspace{-10pt}
\end{figure}

\section{Supplementary Experiments Results}
\label{appendix:more_results}
\paragraph{FTT performs well on Dominoes.}
We first show the main comparison between ERM and FTT on Dominoes in \Cref{fig:dominoes_erm+ftt}. 
We find that, both methods perform well when $\eta_{core} < \eta_{spu}$.
On the contrary, FTT recovers accuracy when $\eta_{core} > \eta_{spu}$ by $4\%$ in average and $9\%$ at most.

\begin{figure}[!b]
    \centering
    \includegraphics[width=0.9\textwidth]{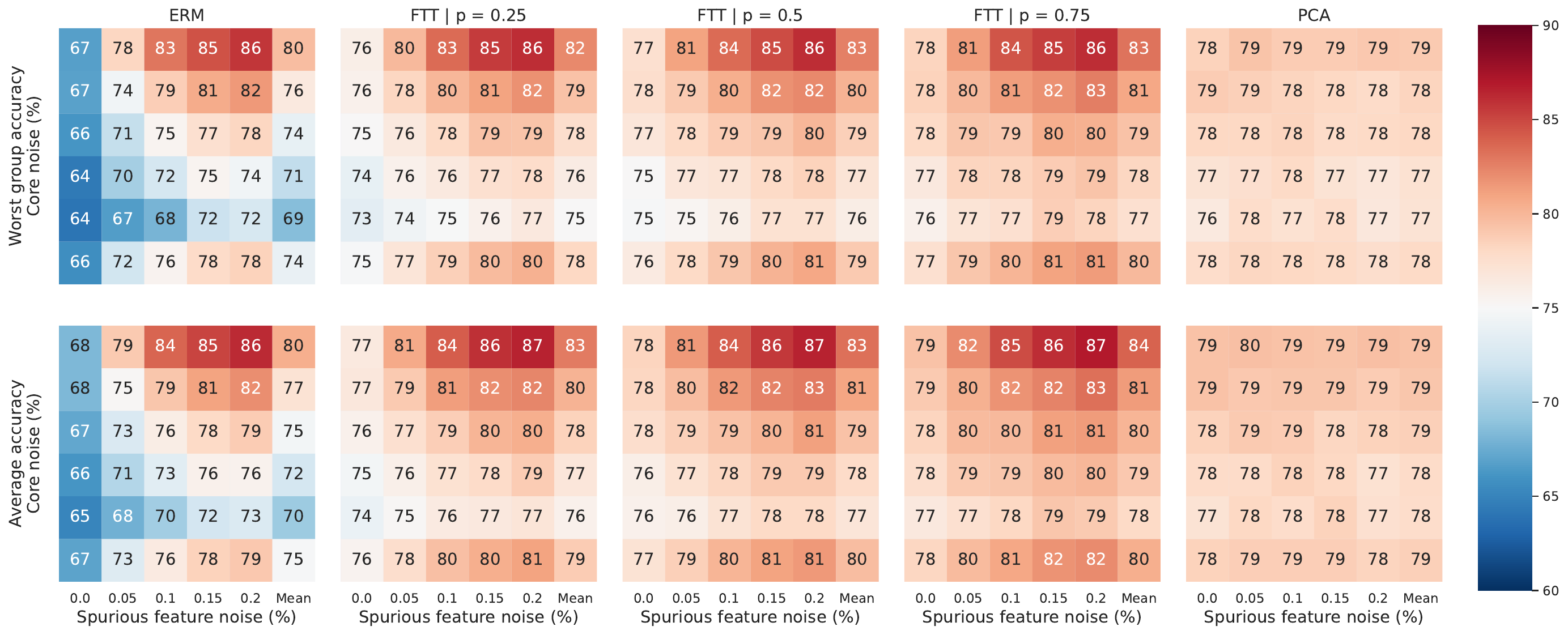}
    \caption{Worst group accuracy (top) and average accuracy (down) on Dominoes for FTT with different $p$. The first and the second column are the same as in \cref{fig:dominoes_erm+ftt}.}
    \label{fig:pvalue_dominoes}
\end{figure}

\paragraph{Experiments on $p$.}
We next show the complete experiments on the unsupervised fraction $p$.
For Waterbirds and CelebA, the results are in \Cref{tab:p_complete}.
We can see that both for worst group accuracy and for average accuracy, both on Waterbirds and on CelebA, FTT performs well under various selections of $p$.
For Dominoes, the results are in \cref{fig:dominoes_erm+ftt}. Again, this plot verifies the effectiveness of FTT, and show that FTT can perform well under different selections of $p$.

\begin{table*}[th]
% \centering
\begin{center}
% \small
\begin{tabular}{ccccccccccccc}
\hline
\multirow{2}{*}{Dataset} & $p$ & 0.0 & 0.25 & 0.5 & 0.75 & 1.0 & \quad &  0.0 & 0.25 & 0.5 & 0.75 & 1.0\\
% \multirow{2}{*}{Method}  & \multirow{2}{*}{Average (\%)} & \multicolumn{4}{c}{Noise Gap (Core - Spurious)} \\
% \cmidrule(r){2-2}  %\cmidrule(r){3-7} \cmidrule(r){9-13}
% \cmidrule(r){2}
\cline{2-2} \cline{3-7} \cline{9-13}
& $\eta_{core}$ & \multicolumn{5}{c}{Worst Group Accuracy (\%)} & & \multicolumn{5}{c}{Average Accuracy (\%)} \\

% &&ERM&CVaR DRO &JTT&Ours & &ERM&CVaR DRO&JTT&Ours\\
\hline
\multirow{7}{*}{Waterbirds}
& 0 & 95.0 & 94.5 & 94.8 & 94.6 & 93.2 &  & 95.3 & 94.9 & 95.2 & 95.0 & 93.7 \\
& 2 & 93.6 & 93.6 & 94.3 & 94.1 & 92.9 &  & 94.2 & 94.2 & 94.5 & 94.6 & 93.5 \\
& 4 & 92.8 & 92.9 & 93.2 & 93.7 & 92.6 &  & 93.2 & 93.5 & 93.6 & 94.0 & 93.0 \\
& 6 & 90.8 & 92.8 & 93.3 & 93.3 & 93.0 &  & 91.3 & 92.9 & 93.3 & 93.4 & 93.4 \\
& 8 & 88.5 & 92.7 & 92.6 & 93.0 & 92.7 &  & 89.9 & 93.0 & 92.9 & 93.3 & 93.1 \\
& 10 & 87.6 & 92.4 & 93.1 & 93.0 & 93.1 &  & 89.4 & 92.9 & 93.5 & 93.3 & 93.4 \\
\cline{2-13}
& Mean & 91.4 & 93.1 & 93.6 & 93.6 & 92.9 &  & 92.2 & 93.6 & 93.8 & 93.9 & 93.4 \\
\hline
\multirow{7}{*}{CelebA} 
& 0 & 95.0 & 95.3 & 95.1 & 95.2 & 92.5 &  & 97.2 & 97.2 & 97.2 & 97.2 & 95.9 \\
& 2 & 95.2 & 95.2 & 95.2 & 94.8 & 92.5 &  & 97.2 & 97.2 & 97.3 & 97.0 & 95.8 \\
& 4 & 94.5 & 94.9 & 94.4 & 94.0 & 92.0 &  & 97.1 & 97.2 & 97.0 & 96.7 & 95.6 \\
& 6 & 94.3 & 94.4 & 94.1 & 94.1 & 92.2 &  & 96.9 & 97.0 & 96.9 & 96.8 & 95.7 \\
& 8 & 93.7 & 94.0 & 93.7 & 93.5 & 92.3 &  & 96.7 & 96.7 & 96.7 & 96.5 & 95.8 \\
& 10 & 92.4 & 93.1 & 92.7 & 93.1 & 91.9 &  & 96.2 & 96.3 & 96.1 & 96.2 & 95.5 \\
\cline{2-13}
& Mean & 94.2 & 94.5 & 94.2 & 94.1 & 92.2 &  & 96.9 & 96.9 & 96.9 & 96.7 & 95.7 \\
\hline
\end{tabular}
\end{center}
\caption{
FTT probing performance on Waterbirds and CelebA, under different unsupervised fraction $p$. 
This table is an extension of \cref{tab:p_value}, which only contains the worst group accuracy on Waterbirds.
}
\label{tab:p_complete}
\end{table*}

\section{Proof of Theorems}
\label{appendix:proofs}
For simplicity, for all the proofs below, we rewrite $\epsilon_{core}, \epsilon_{spu}$ as $\epsilon_1, \epsilon_2$, $\eta_{core}, \eta_{spu}$ as $\eta_1, \eta_2$, and $\mW_{core}, \mW_{spu}$ as $\mW_1, \mW_2$.
Denote the covariance matrix of $\vx$ as $\mH = \mathbb E[\vx^\top \vx]$ (notice that $\vx \in \mathbb R^{d \times 1}$).
By standard algebra, we have
\begin{align}
    \mH = \begin{pmatrix}
        \mSigma & \mSigma \beta \gamma^\top \\ \gamma \beta^\top \mSigma & \eta_2^2 \mI + (\eta_1^2 + \beta^\top \mSigma \beta)\gamma \gamma^\top
    \end{pmatrix}.
\end{align}
We denote the SVD decomposition of $\mH = \Xi \mD \Xi^\top$, where $\Xi$ is an orthogonal matrix and $\mD$ is a diagonal matrix in descending order.
We denote the SVD decomposition of $\mSigma = \mQ \mD_1 \mQ^\top$, where $\mQ = (\vq_1,\cdots, \vq_{d_1})$ is an orthogonal matrix and $\mD_1 = \text{Diag}\left( \lambda_1,\cdots, \lambda_{d_1} \right)$ is a descending diagonal matrix.
We use $\mA_{:n}$ to denote the first $n$ columns of $\mA$, and $\text{span}(\mA)$ to denote the linear space spanned by the column vectors of $\mA$.
% Recall that $\beta \in \text{span}(\mV_{:n}) \subset \text{span}(\mV_{:pm})$, 
Denote $g = \beta^\top \mSigma \beta$ as the variance of $\vx_1$ along the ground-truth direction $\beta$. 
Also recall that $\alpha = \frac{\eta_2^2}{\eta_1^2 + \eta_2^2}$.

Notice that $\beta$ lies in the top $k$ eigenvectors spanned space, i.e. $\beta \in \text{span} (\mQ_{:k})$.
Without loss of generality, we assume that $k$ is the minimum integer that satisfies this condition, i.e. we assume that $\vq_k^\top \beta \not= 0$.
Otherwise, we decrease $k$ until this is true, while the condition $p > \frac k m$ still holds.

\input{Appendix/upper_bound.tex}

\input{Appendix/lower_bound.tex}

\input{Appendix/FTT_bound.tex}

\input{Appendix/Proof_of_lemmas.tex}

%% file: Appendix/upper_bound.tex
\subsection{Proof of \cref{theorem_upper}}
\label{appendix:upper}

\textbf{Proof sketch.} To upper bound $\ell_{te}(\mW(t))$, we demonstrate that by a proper selection of $\vb$, we have $\ell_{te}(\mW(t)) \leq \frac{1}{(1-\gamma^\top \mW_2(t) \vb(t))^2} \ell_{tr}(\mW(t), \vb(t))$ (\Cref{lemma_upper_decomposion}). 
Since $\eta_{core} < \eta_{spu}$, we can prove that the weight assigned on $\vx_2$ is upper bounded, which help control the magnitude of $\gamma^\top \mW_2(t) \vb(t)$. 
On the other hand, we follow the idea from \cite{ali2019continuous} to control $\ell_{tr}(\mW(t), \vb(t))$ using some differential equation techniques (\Cref{lemma:bound_training_error}).
This helps circumvent the direct analysis on the not-close form solution.

To upper bound test error, we first connect it with training error using the following lemma.
\begin{lemma}
\label{lemma_upper_decomposion}
    For all $ \mW, \vb$, we have
    \begin{align}
        \ell_{te}(\mW) \leq \frac{1}{(1 - \gamma^\top \mW_2 \vb)^2}\ell_{tr}(\mW ,\vb).
    \end{align}
\end{lemma}

\Cref{lemma_upper_decomposion} decomposes $\ell_{te}(\mW(t))$ into a factor and the training error. For simplicity, we denote $c_t = \gamma^\top \mW_2(t) \vb(t)$. Below we separately bound both terms.

\paragraph{Bound $\ell_{tr}(t)$.} 
\cite{ali2019continuous} has pointed out that continuous time linear regression (i.e. one layer network) gives an analytical solution $\vv(t)$. 
For a two layer model this remains unknown.
% However, below we show that even an analytical solution is unspecified, we can still bound the error similarly. 
However, since this optimization problem is convex in terms of $\vv$ (but not of $(\mW,\vb)$), in the infinity $\ell_{tr}(t)$ we can still bound the training error, which is specified by the following lemma.

\begin{lemma}
\label{lemma:bound_training_error}
    Under the assumption in \Cref{theorem_upper}, for all time step $t$, the training error is bounded by
    \begin{align*}
        \ell_{tr}(\mW(t) ,\vb(t))  \leq {\rm err}_{tr}^* + \mathcal O\left(\frac 1 t \right).
    \end{align*}
\end{lemma}

\paragraph{Bound $c_t$.}
Notice that by standard decomposition, we have
$$
\ell_{tr}(\mW(t),\vb(t)) \geq \frac 1 2 \left( (1-c_t)^2 \eta_1^2 + \eta_2^2 \|\vv_2(t)\|^2_2 \right) 
\geq \frac 1 2 \left( (1-c_t)^2 \eta_1^2 + \eta_2^2 c_t^2 \right).
$$
Therefore, for all time step $t$, we have
$$
\frac{\eta_1^2 - \sqrt{2\ell_{tr} (\eta_1^2 + \eta_2^2) - \eta_1^2 \eta_2^2}}{\eta_1^2 +\eta_2^2}\leq c \leq  \frac{\eta_1^2 + \sqrt{2\ell_{tr} (\eta_1^2 + \eta_2^2) - \eta_1^2 \eta_2^2}}{\eta_1^2 +\eta_2^2}.
$$

Together, we have
\begin{align*}
    \ell_{te}(\mW(t)) 
    &\leq \frac 1 {(1-c_t)^2} \ell_{tr}(\mW(t), \vb(t)) \\
    &\leq \frac 1 {(1-c_t)^2} \left(\mathcal{O} (\frac 1 {4t}) + \text{err}_{tr}^*\right) \\
    &\leq \left(\frac{\eta_1^2 + \eta_2^2}{\eta_2^2 - \sqrt{2\ell_{tr}(\mW(t), \vb(t))(\eta_1^2 + \eta_2^2) - \eta_1^2 \eta_2^2 }}\right)^2\left(\mathcal{O} (\frac 1 {4t}) + \text{err}_{tr}^*\right) \\
    &\leq \left(\frac{\eta_1^2 + \eta_2^2}{\eta_2^2 - \mathcal O (t^{-1/2})
    }\right)^2\text{err}_{tr}^* + \mathcal{O} (t^{-1})  \\
    & = \left(1 + \frac{\eta_1^2}{\eta_2^2}\right) \text{err}^*_{te} + \mathcal O(t^{-1}).
\end{align*}

%% file: Appendix/lower_bound.tex
\subsection{Proof of \Cref{theorem_lower}}
\label{appendix:lower}

\textbf{Proof sketch of \Cref{theorem_lower}.} 
To prove the theorem, we first analyze the optimal selection of $\vb$ given a feature matrix $\mW$.
We then convert the test error to a expression that depends on the norm of $\mW$ in the infinity (\Cref{lemma:lower_simplify_error}). 
We then leverage the fact that
$$
\partial_t \left(\mW(t)^\top \mW(t) - \vb(t) \vb(t)^\top \right) = \mathbf 0
$$
to connect the parameters across different $t$, and control the matrix norm using the properties of initialization (\Cref{lemma:lower_2norm}).

We already know that $\lim_{t\to\infty}\mW(t)\vb(t) = \mW(\infty)\vb(\infty) = \vv^*$. Since $\mW(\infty)$ has full column rank, exists $ T_0$ such that for all $t > T_0$, $\mW(t)$ has full rank and the $m^{th}$ singular value is simultaneously lower bounded by a positive constant $\lambda_0$ that depends only on $\mW(\infty)$.

For any fix $\mW$, and the test error for a given $\vb$ is
\begin{align}
\label{eq:test_error_fix}
    \mathbb E_{\vx_2 = \epsilon_2} \frac 1 2 \|\vx \mW \vb - y \|^2 
    &= \frac 1 2 \left( \mathbb E [\vx_1 \mW_1 \vb - y]^2 + \mathbb E [\vx_2 \mW_2 \vb]^2 + \mathbb E [\vx_1 \mW_1\vb\cdot \vx_2 \mW_2 \vb]  \right)\\
    &= \frac 1 2\left( \eta_1^2 + \mathbb E\| \mW_1 \vb - \beta\|_\mSigma^2 + \eta_2^2 \| \mW_2 \vb \|^2_2  \right).
\end{align}
Since \Cref{eq:test_error_fix} is convex w.r.t. $\vb$, it is minimized when
\begin{align}
    \mathbf 0 = \nabla_\vb \mathbb E_{\vx_2 = \epsilon_2} \frac 1 2 \|\vx \mW \vb - y \|^2 =   \left(\mW_1^\top \mSigma \mW_1 + \eta_2^2 \mW_2^\top \mW_2\right)\vb - \mW_1^\top \mSigma \beta
\end{align}
The quadratic form will not degenerate so long as $t > T_0$ and $\mW_1(t)^\top \mSigma \mW_1(t)$ is positive definite.
Together, the test error is minimized in time $t$ by setting
\begin{align*}
    \vb_{\min}(t) = \left(\mW_1^\top(t) \mSigma \mW_1(t) + \eta_2^2 \mW_2^\top(t) \mW_2(t)\right)^{-1} \mW_1^\top(t) \mSigma \beta.
\end{align*}
For all $t > T_0$, $\vb_{\min}(t)$ is continuous in terms of $t$, i.e.
$$
\vb_{\min}(t) \to \vb_{\min}(\infty) =  \left(\mW_1^\top(\infty) \mSigma \mW_1(\infty) + \eta_2^2 \mW_2^\top(\infty) \mW_2(\infty)\right)^{-1} \mW_1^\top(\infty) \mSigma \beta.
$$
Therefore, $\lim_{t\to \infty}\ell_{te}(\mW(t)) = \ell_{te}(\mW(\infty))$, and the latter is minimized by setting $\vb$ to $\vb_{\min}(\infty)$.
For simplicity, we abbreviate $\mW_1(\infty), \mW_2(\infty)$ as $\mW_1,\mW_2$.
\Cref{lemma:lower_simplify_error} below helps simplify the infinite error term, and \Cref{lemma:lower_2norm} helps bound the simplified error.

\begin{lemma}
    \label{lemma:lower_simplify_error}
    Under the condition in \Cref{theorem_lower}, we have
    \begin{align}
        \frac{\ell_{te}(\mW)}{\text{err}^*_{te}} = 1 + \frac{\eta_1^2}{\eta_2^2}\gamma^\top \left(\mI + \eta_2^2 \mW_2\left[ \mW_1^\top \mSigma \mW_1
 \right]^{-1}\mW_2^\top \right)^{-1}\gamma
    \end{align}
\end{lemma}

\begin{lemma}
\label{lemma:lower_2norm}
    Under the Xavier uniform initialization, 
    \begin{align}
    \lambda_{\max}^{-1} \left(\mI + \eta_2^2 \mW_2\left[ \mW_1^\top \mSigma \mW_1
 \right]^{-1}\mW_2^\top \right) \geq \frac 12 \wedge	 \frac 1 {4\eta_2^2 \|\mSigma\|_2 \|\mW_1^+\|^2_2}.
\end{align}
\end{lemma}

Combining \Cref{lemma:lower_simplify_error,lemma:lower_2norm}, we have
\begin{align*}
    \lim_{t\infty} \frac{\ell_{te}(\mW(t)}{\text{err}^*_{te}} 
    &= \frac{\ell_{te}(\mW(\infty))}{\text{err}^*_{te}} \\
    &= 1+ \frac{\eta_1^2}{\eta_2^2} \gamma^\top \left(\mI + \eta_2^2 \mW_2\left[ \mW_1^\top \mSigma \mW_1 \right]^{-1}\mW_2^\top \right)^{-1}\gamma \\
    &\geq 1 + \frac{\eta_1^2}{\eta_2^2} \lambda_{\max}^{-1}\left(\mI + \eta_2^2 \mW_2\left[ \mW_1^\top \mSigma \mW_1 \right]^{-1}\mW_2^\top  \right) \|\gamma\|^2 \\
    &\geq 1 + \frac{\eta_1^2}{\eta_2^2} \left(\frac 12 \wedge  \frac 1 {4\eta_2^2\|\mSigma\|_2 \|\mW_1^+(\infty)\|_2^2} \right).
\end{align*}

%% file: Appendix/FTT_bound.tex
\subsection{Proof of Theorem~\ref{theorem_FTT}}
\textbf{Proof sketch.}
The key intuition is that $\mW_{ul}$ recovers important information about $\beta$, despite there is error in PCA that we can never full recover $\beta$.
In this case, we can combine the features learned in $\mW_{sl}$ and $\mW_{ul}$ to obtain a asymptotically optimal approximation of $\beta$ without being disturbed by the spurious correlation $\gamma$.
Specifically, we prove the following lemma.

\begin{lemma}
    \label{lemma:span_of_ul}
    Exists $c_1 \not = \frac{1-\alpha}{\alpha}$, such that
    \begin{align}
         \begin{pmatrix}
            \beta \\ c_1 \gamma
        \end{pmatrix}
        \in \text{span}\left(\Xi_{:k}\right).
    \end{align}
\end{lemma}

Since our unsupervised training features $\mW_{ul}$ takes the top $pm$ eigenvectors of $\Xi$, \cref{lemma:span_of_ul} implies that $\left(\beta^\top, c_1 \gamma^\top\right)^\top$ lies in the span of $\mW_{ul}$, i.e. $\exists \hat \vb_{ul} \in \mathbb R^{pm \times 1}$, such that $\left(\beta^\top, c_1 \gamma^\top\right)^\top = \mW_{ul} \hat \vb_{ul}$.
On the other hand, we alredy know from the proof of \cref{theorem_upper} in the infinity $\vv(t) \to \left(\alpha\beta^\top, (1-\alpha)\gamma^\top \right)^\top$.
The following lemma, by combing these two crucial feature, bounds the test time probing error of $\mW_{FTT}(t)$.

\newcommand{\scalete}{{\frac{c_1}{c_1 \alpha - (1-\alpha)}}}

\begin{lemma}
    \label{lemma:FTT_error}
    By setting the retraining weight 
    \begin{align*}
        \hat \vb = \scalete\mW_{FTT}(t)\vb{(t)} - \frac{1-\alpha}{c_1\alpha - (1-\alpha)} \begin{pmatrix}
            \hat  \vb_{ul} \\ \mathbf 0
        \end{pmatrix},
    \end{align*}
    we have (recall that $c_t = \gamma^\top \mW_2(t) \vb(t)$, while $c_1$ is fixed)
    \begin{align}
        \ell_{te}(\mW_{FTT}(t)) - \text{err}_{te}^* 
        &\leq \left(\scalete \right)^2 \mathcal O \left( \frac {\|\mSigma\|_2 \|\mH^{-1}\|_2} {(\sqrt{c_0^2 + 1} - 1)t}\right).
    \end{align}
    Here $\mathcal O$ hides a universal constant.
\end{lemma}
Notice that the RHS decays with rate $t^{-1}$. Together, the proof is finished.

%% file: Appendix/Proof_of_lemmas.tex
\section{Proof of Lemmas}
In this section we prove all lemmas.

\subsection{Proof of Lemma~\ref{lemma_optimal_training_error}}
According to the decomposition of the training error, for any $\vv = \mW\vb$, we have
\begin{align*}
    \ell_{tr}(\mW, \vb) 
    &= \frac 1 2 \left( \mathbb E \|\vx_1 \vv_1 - (1-\gamma^\top \vv_2)y\|^2 + \eta_2^2 \|\vv_2\|^2 \right) \\
    &= \frac 1 2 \left( \mathbb E \|\vx_1 \vv_1 - (1-\gamma^\top \vv_2)\vx_1 \beta\|^2 + \eta_1^2(1-\gamma^\top \vv_2)^2 + \eta_2^2 \|\vv_2\|^2 \right) \\ 
    &\geq \frac 12 \left( \eta_1^2(1-\gamma^\top \vv_2)^2 + \eta_2^2 \|\vv_2\|^2 \right) 
\end{align*}
Denote $x = \gamma^\top \vv_2$, we have
$$
\ell_{tr}(\mW, \vb) \geq \frac 1 2 \left(\eta_1^2 (1-x)^2 + \eta_2^2 x^2 \right) \geq \frac 1 2 \cdot \frac{\eta_1^2\eta_2^2}{\eta_1^2 + \eta_2^2}.
$$
Here the last inequality comes from Cauchy–Schwarz inequality.
The proof is finished by verifying that $\vv^*_{tr}$ can indeed give the minimum.

\subsection{Proof of Lemma~\ref{lemma_upper_decomposion}}
Given $\mW, \vb$, we can decompose the training error as (denote $\vv = \mW \vb$)
\begin{align*}
    \ell_{tr}(\mW, \vb) 
    &= \frac 1 2 \mathbb E \|\vx_1 \vv_1  + (y\gamma^\top +\epsilon_2) \vv_2 - y \|_2^2 \\
    &=  \frac 1 2\left( \mathbb E \|\vx_1 \vv_1 - (1-\gamma^\top \vv_2) y \|^2 + \eta_2^2 \|\vv_2\|^2\right) \\
    % &= \frac 1 2\left( \mathbb E \| \vx_1 \vv_1 - (1-\gamma^\top \vv_2) \vx_1\beta \|^2 + (1 - \gamma^\top \vv_2)^2 \eta_1^2 + \eta_2^2 \|\vv_2\|^2 \right)\\
    % &= \frac 1 2\left( \|\vv_1 - (1-\gamma^\top \vv_2)\beta \|_\mSigma^2 +  (1 - \gamma^\top \vv_2)^2 \eta_1^2 + \eta_2^2 \|\vv_2\|^2  \right)
\end{align*}
On the other hand, by setting $\hat \vb = \frac 1 {1 - \gamma^\top \vv_2}\vb$, the test error is upper bounded by
\begin{align*}
    \ell_{te}(\mW) 
    &\leq \frac 1 2 \mathbb E_{\vx_2 = \epsilon_2} \|\vx \mW \hat \vb - y\|^2 \\
    &= \frac 1 2 \mathbb E_{\vx_2 = \epsilon_2} \|\frac {\vx \vv}{1-\gamma^\top\vv_2} - y\|^2 \\
    &= \frac 1 2 \left(\mathbb E \|\frac {\vx_1 \vv_1}{1-\gamma^\top\vv_2} - y\|^2 + \frac {\eta_2^2}{(1-\gamma^\top \vv_2)^2}\|\vv_2\|^2 \right) \\
\end{align*}
Therefore, we have
\begin{align*}
    \ell_{te}(\mW) \leq \frac{1}{(1 - \gamma^\top \mW_2 \vb)^2}\ell_{tr}(\mW ,\vb).
\end{align*}

\subsection{Proof of Lemma~\ref{lemma:bound_training_error}}
Denote $\mM(t) = \mW(t)^\top \mW(t) - \vb(t)\vb(t)^\top $. 
Since our parameters are initialized according to Xaiver uniform distribution, 
$$
\|\mM(0)\|_2 \leq \|\mW(0)\|^2_2 + \|\vb(0)\|_2^2 \leq d \|\mW(0)\|_\infty^2 + 1 \leq 2.
$$
Recall that $\mM(t)$ is invariant throughout the whole optimization.
We have
$$
2\|\vb(t)\|_2^2 \geq \vb(t)^\top \mM(0) \vb(t) = \vb(t)^\top \mM(t) \vb(t) = \|\vv(t)\|_2^2 - \|\vb(t)\|^4,
$$
which implies that $\|\vb(t)\|_2^2 \geq \sqrt{c_0^2 + 1} - 1$.

\paragraph{Convergence of $\vv(t)$.}The gradient of $\vv(t)$ is 
\begin{align*}
    \partial \vv(t) 
    &= \partial_t \mW(t) \cdot \vb(t) + \mW(t) \cdot \partial_t \vb(t) \\
    &= \left(\mW(t)\mW(t)^\top + \|\vb(t)\|_2^2 \mI \right) \left( \mathbb E[\vx^\top y] - \mH \vv(t)\right) \\
    &= \left(\mW(t)\mW(t)^\top + \|\vb(t)\|_2^2 \mI \right)\mH \left( \vv_{tr}^* -  \vv(t)\right)
\end{align*}
where the last equation uses the fact that $\mH$ is invertible (since $\mSigma$ is invertible) and $\vv_{tr}^* = \mH ^{-1} \mathbb E[\vx^\top y]$. 
By a standard differential equation analysis, we have
\begin{align}
\label{eq:ode}
    \vv(t) - \vv_{tr}^* = \exp\left\{ 
-\mH \int_0^t \mA(\tau) \mathrm d\tau \right\} (\vv(0) - \vv_{tr}^*),
\end{align}
where $\mA(t) \triangleq \mW(t)\mW(t)^\top + \|\vb(t)\|_2^2 \mI$ and $\mA(t) - (\sqrt{c_0^2 + 1} -1)\mI$ is positive definite because the bound of $\|\vb(t)\|^2_2$.
This help us control the training error as
\begin{align*}
    2\ell_{tr}(\mW(t), \vb(t)) 
    &=  \mathbb E \|\vx \vv(t) - y \|_2^2 \\
    &= \vv(t)^\top \mH \vv(t) - 2 \vv(t)^\top \mathbb E[\vx^\top y] + \mathbb E [y^\top y]\\
    &= \vv(t)^\top \mH \vv(t) - 2\vv(t)^\top \mH \vv_{tr}^* + \vv_{tr}^{*\top} \mH \vv_{tr}^* \\
    &\quad + \left( \vv_{tr}^{*\top} \mH \vv_{tr}^* - 2\vv_{tr}^{*\top} \mH \mathbb E[\vx^\top y] + \mathbb E[y^\top y\right) \\
    &= \| \vv(t) - \vv_{tr}^* \|_{\mH}^2 + \mathbb E \|\vx \vv_{tr}^* - y\|_2^2 \\
    &= \| \vv(t) - \vv_{tr}^* \|_{\mH}^2 + 2\text{err}_{tr}^*.
\end{align*}
Plugging \Cref{eq:ode} into the first term, we have
\begin{align*}
    \| \vv(t) - \vv_{tr}^* \|_{\mH}^2 
    &= (\vv(0) - \vv_{tr}^*)^\top \left[ \exp\left\{ 
-2\mH \int_0^t \mA(\tau) \mathrm d\tau \right\} \mH
 \right] (\vv(0) - \vv_{tr}^*) \\
 &\leq \mathcal O\left(\left\| \exp\left\{ 
-2\mH \int_0^t \mA(\tau) \mathrm d\tau \right\} \mH \right\|_2\right) \\
&\leq \mathcal O\left(\left\| \exp\left\{ 
-2 (\sqrt{c_0^2 + 1 } -1 )t\mH  \mathrm d\tau \right\} \mH \right\|_2\right) \\
&\leq \mathcal O \left( \frac 1 {(\sqrt{c_0^2 + 1} - 1)t}\right).
\end{align*}
Here the first equation is because $\mH$ and $\int_0^t \mA(\tau)$ are both positive definite, and can be diagonalized simultaneously. The first inequality is because $\|\vv(0) - \vv_{tr}^*\|$ is bounded, while the second is because $\mA(t) - (\sqrt{c_0^2+1} - 1) \mI$ is positive definite for all $t$.

\subsection{Proof of Lemma~\ref{lemma:lower_simplify_error}}
Plug in $\vb = \vb_{\min}(\infty)$ into \Cref{eq:test_error_fix}, we obtain (denote $\mLambda = \mW_1^\top \mSigma \mW_1$)
\begin{align*}
\ell_{te}(\mW_1) - \text{err}^*_{te}  
&= \frac 1 2 \left( \vb^\top \left[\mLambda + \eta_2^2 \mW_2^\top \mW_2 \right]^{-1} \vb - 2\beta^\top \mSigma \mW_1 \vb + \beta^\top \mSigma \beta \right) \\
&= \frac 1 2 \left(\beta^\top \mSigma \beta - \beta^\top \mSigma \mW_1 \left[\mLambda + \eta_2^2 \mW_2^\top \mW_2 \right]^{-1} \mW_1^\top \mSigma \beta \right) \\
&= \frac 1 2 \left(\beta^\top \mSigma \beta - \beta^\top \mSigma \mW_1 \left[\mLambda^{-1} - \eta_2^2 \mLambda^{-1} \mW_2 \left(\mI + \eta_2^2 \mW \mLambda^{-1} \mW_2^\top \right)^{-1} \mW_2 \mLambda^{-1} \right] \mW_1^\top \mSigma \beta \right)
\end{align*}
where the last equation is because for any invertible $\mA,\mB$,
$$
\left(\mA + \mC \mB \mC^\top \right)^{-1} = \mA^{-1} - \mA^{-1} \mC \left( \mB^{-1} + \mC^\top \mA^{-1}\mC \right)^{-1}\mC^\top \mA^{-1}.
$$

Notice that we have $\mW_1 \vb(\infty) = \alpha \beta, \mW_2 \vb(\infty) = (1-\alpha) \gamma$. 
Multiplying $\mW_1^\top \mSigma$ on both side, we have
\begin{align*}
\mLambda \vb(\infty) &= \alpha \mW_1^\top \mSigma \beta \\
\vb(\infty) &= \alpha \mLambda^{-1} \mW_1^\top \mSigma \beta.
\end{align*}
This implies that
\begin{align*}
    \beta^\top \mSigma \mW_1 \mLambda^{-1} \mW_1^\top \mSigma \beta &= \frac 1 {\alpha} \beta^\top \mSigma \mW_1 \vb(\infty) = \beta^\top \mSigma \beta, \\
    \mW_2 \mLambda^{-1} \mW_1^\top \mSigma \beta &= \frac 1 \alpha \mW_2 \vb(\infty) = \frac{1-\alpha}{\alpha} \gamma. 
\end{align*}
Therefore, we have
\begin{align*}
\ell_{te}(\mW_1) - \text{err}^*_{te}  
&= \frac {(1-\alpha)^2} {2\eta_2^2 \alpha^2} \left(  \gamma^\top \left[\mI + \eta_2^2 \mW \mLambda^{-1} \mW_2^\top \right]^{-1} \gamma \right)\\
&= \frac{\eta_1^4}{2\eta_2^2} \gamma^\top \left[\mI + \eta_2^2 \mW \mLambda^{-1} \mW_2^\top \right]^{-1} \gamma.
\end{align*}

\subsection{Proof of Lemma~\ref{lemma:lower_2norm}}
We first upper bound $\|\mW_2\|_2$. Recall an important property of our two layer linear model from \cite{kumar2022fine}:
\begin{align}
    \partial_t \left[ \mW(t)^\top \mW(t)^\top - \vb(t) \vb(t)^\top \right] = \mathbf 0.
\end{align}
Applying this property with $t = 0$ and $t\to \infty$, we have
$$
\mW_1 ^\top \mW_1 + \mW_2 ^\top \mW_2 - \vb(\infty) \vb(\infty)^\top = \mW(0)^\top \mW(0) - \vb(0) \vb(0)^\top.
$$
By multiplying $\mW_2$ on the left and $\mW_2^\top$ on the right,
$$
\lambda_{\max}\left(\mW_2 \mW_2^\top \mW_2 \mW_2^\top - (1-\alpha)^2 \gamma \gamma^\top  \right) \leq \lambda_{\max}\left(\mW_2 \mW(0)^\top \mW(0) \mW_2^\top\right),
$$
which implies that
$$
\|\mW_2\|_2^4 - 1 \leq \| \mW_2\|^2 \| \mW(0)\|_2^2.
$$
Since $\mW(0)$ is initialized according to the Xavier uniform distribution, we have
$$
\|\mW(0) \| \leq \sqrt{d} \|\mW(0)\|_\infty = \sqrt d \cdot \frac 1 {\sqrt{d}} = 1.
$$
This quickly implies that $\|\mW_2\|_2^2 < 2$.

Together, we have
\begin{align*}
    \lambda_{\max} \left(\mI + \eta_2^2 \mW_2\left[ \mW_1^\top \mSigma \mW_1
 \right]^{-1}\mW_2^\top \right)
 &\leq 1 + \eta_2^2 \|\mW_2\|_2^2 \lambda_{\max}\left( \left[ \mW_1^\top \mSigma \mW_1 \right]^{-1} \right) \\
 &\leq 1 + 2 \eta_2^2  \lambda_{\max}\left( \left[ \mW_1^\top \mSigma \mW_1 \right]^{-1} \right) \\
 &\leq 1 + 2 \eta_2^2 \|\mSigma^{-1}\|_2 \|\mW_1^+\|_2^2\\
 &\leq 2\max\left\{1, 2 \eta_2^2 \|\mSigma^{-1}\|_2 \|\mW_1^+\|_2^2\right\}
\end{align*}
where the second last inequality comes from standard linear algebra. The proof is finished by taking the inverse.

\subsection{Proof of Lemma~\ref{lemma:span_of_ul}}
To study the span of $\Xi_{:k}$, we first need to understand the property of the eigenvector of $\mH$. Notice that $\mH$ is p.d. and all eigenvalue is positive.
The eigenvectors of $\mH$ can be divided into the following three groups.

\begin{enumerate}
    \item First, consider any normalized vector $\gamma^\perp \in \mathbb R^{d_2 \times 1}$. In this case, we have
    $$
    \mH \begin{pmatrix}
        \mathbf 0 \\ \gamma^\perp
    \end{pmatrix}
    = \eta_2^2 \begin{pmatrix}
        \mathbf 0 \\ \gamma^\perp
    \end{pmatrix}.
    $$
    Since the space dimension of \{$\gamma^\perp\}$ is $d_2 - 1$, we find $d_2-1$ eigenvector of $\mH$ with eigenvalue $\eta_2^2$. We denote them as $(\mathbf 0^\top, \gamma^\perp_i)^\top, i \in [d_2-1]$.
    \item Second, for $j > k$,
    $$
    \mH \begin{pmatrix}
        \vq_j \\ \mathbf 0
    \end{pmatrix}
    = \begin{pmatrix}
        \mSigma \vq_j \\ \beta^\top \mSigma \vq_j 
    \end{pmatrix} 
    = \lambda_j \begin{pmatrix}
        \vq_j \\ \mathbf 0 
    \end{pmatrix}
    $$
    since $\beta$ lies in the span of the top $k$ eigenvectors of $\mSigma$. This implies that for $j = k+1,\cdots, d_1$, $(\vq_i^\top,\mathbf 0^\top)^\top$ is the eigenvector of $\mH$ with eigenvalue $\lambda_i$.
    
    \item All the eigenvector left. We denote them as $\vu_i$ with eigenvector $\mu_i$ for $i \in [k + 1]$. Notice that $\vq_i (i\leq k)$ is not an eigenvector so long as $\beta^\top \vq_i \not= 0 $.
\end{enumerate}

The essential intuition is that for at least $k$ vectors in group 3, their eigenvalue is strictly larger than all eigenvalues in group 1 ($\eta_2^2$) and group 2 ($\lambda_j, j > k$). To see this, denote $\vu_i = (\vu_{i,1}^\top, \vu_{i,2}^\top)^\top$.
First, notice that $\vu_{i,2} \perp \gamma^\perp$. 
Therefore, $\vu_{i,2}$ must be in the direction of $\gamma$, i.e. $\vu_{i,2} = r_i \gamma$, where $r_i$ can possibly be $0$. 
Second, $\vu_{i,1} \perp \vq_j, \forall j > k$.
Therefore, we can further denote $\vu_{i,1} = \sum_{\tau = 1}^k e_{i,\tau} \vq_\tau$.
In this case, since
$$
\mu_i \begin{pmatrix}
    \vu_{i,1} \\ r_i \gamma
\end{pmatrix}
    =\mH \vu = \begin{pmatrix}
    \mSigma \vu_{i,1} + r_i \mSigma \beta  \\
    \beta^\top \mSigma \vu_{i,1} + r_i (\eta_1 + \eta_2 + g) \gamma 
\end{pmatrix},
$$
which implies that for all $i \in [k+1]$,
\begin{align}
\label{eq:eigvalue1}
    e_{i,\tau} \mu_i &= e_{i,\tau} \lambda_{\tau} + r_i\lambda_{\tau} b_\tau, \forall \tau \in [k]\\
\label{eq:eigvalue2}
    \mu_i r_i &= \sum_{\tau = 1}^k \lambda_\tau  b_\tau (e_{i,\tau} + r_i b_\tau) + r_i(\eta_1^2 + \eta_2^2) .
\end{align}
Here $\beta = \sum_{\tau = 1}^k  b_{\tau} \vq_\tau$ and $b_{k} \not = 0$.
With this, we now specify all $k+1$ eigenvalues, start from $i = k+1$ to $i=1$.
We will show that they separately fall in the interval $[0, \lambda_k], [\lambda_k, \lambda_{k-1}],\cdots,[\lambda_2,\lambda_1],[\lambda_1,+\infty)$.
% When $r_i = 0$, we set $\mu_i = $

First, given any $i \in [k]$, When $b_i = 0$, we can set $r_i = 0, e_{i} = 1, \mu = \lambda_{i}$ and $e_{i'} = 0, \forall i' \not= i$.
In this case, \cref{eq:eigvalue1,eq:eigvalue2} are satisfied.
We then find an eigenvector $(\vq_i^\top,\mathbf 0^\top)^\top$ with eigenvalue $\lambda_{i}$.

On the other hand, when $b_{i} \not = 0$, there must be $\mu \not= \lambda_{i}$. We set 
$$
e_{\tau} = \frac{r \lambda_{\tau} b_\tau}{\mu - \lambda_\tau}, \forall \tau \in [k],
$$
which satisfies \cref{eq:eigvalue1}. Plugging into \cref{eq:eigvalue2}, we have
\begin{align}
    \mu r = \sum_{\tau = 1}^k \frac{r \mu \lambda_{\tau} b_\tau^2}{\mu - \lambda_{\tau}} + r(\eta_1^2 + \eta_2^2),
\end{align}
which implies (denote $\sT = \{i \in [k]: b_i \not= 0\}$)
\begin{align}
\label{eq:mu}
    \mu = \sum_{\tau \in \sT} \frac{\mu \lambda_{\tau} b_\tau^2}{\mu - \lambda_{\tau}} + (\eta_1^2 + \eta_2^2).
\end{align}
Any positive solution $\mu$ of \cref{eq:mu} can generate an eigenvector with eigenvalue $\mu$.
% , and the number of eigenvectors (geometric multiplicity) is exactly the multiplicity of $\mu$.
On the other hand, in each interval $[\lambda_\tau, \lambda_{\tau'}]$, the RHS decreases as $\mu$ increases, while the LHS increases as $\mu$ increases. In addition, the RHS goes to $+\infty$ when $\mu \to \lambda_{\tau}^+$ and goes to $-\infty$ when $\mu \to \lambda_{\tau'}^-$. Therefore, there will be exactly one solution for all the intervals in $[0,+\infty)$ divided by elements in $\sT$.
Together, $k+1$ eigenvalues are generated, which is exactly the number of eigenvectors that do not belong to group 1 and 2.
Since we have $\lambda_k < \lambda_{k-1}$ (strictly less), only $\mu_{k+1} \leq \lambda_{k}$, while the rest $\mu$ is lower bounded by $\lambda_{k}$.
Finally, notice that $\lambda_{k} > \eta_1^2 + \eta_2^2$ and $\lambda_{k} > \lambda_{j}, \forall j > k$, we conclude that $\vu_1, \cdots, \vu_k$ is the top $k$ eigenvectors, i.e. $\Xi_{:k}$.

\textbf{Important features in $\text{span}(\Xi_{:k})$.}
Notice that both $(\beta^\top, \mathbf 0 ^\top)^\top$ and $(\mathbf 0^\top, \gamma^\top )^\top$ are orthogonal to eigenvectors in group 1 and 2.
As a result, they must be in the span of $\vu_1,\cdots,\vu_{k+1}$, though they are not in the span of $\Xi_{:k}$.
Nevertheless, since $\text{rank}(\Xi_{:k}) = \text{rank}\left(\text{span}\{\vu_1,\cdots,\vu_{k+1}\} \right)-1$, there must exist $c_1$ such that $(\beta^\top, c_1 \gamma^\top)^\top \in \text{span}(\Xi_{:k})$.
Finally, can $c_1 = \frac{1-\alpha}{\alpha}$?
If so, we have
$$
(\beta^\top, c_1 \gamma^\top) \begin{pmatrix}
    \vu_{k+1,1} \\ r_{k+1} \gamma
\end{pmatrix}
= 0,
$$
which implies that $\lambda_{k+1} = \eta_2^2$. Plugging into \cref{eq:mu}, we have
$$
-\eta_1^2 = \sum_{\tau = 1}^k \frac{\eta_2^2 \lambda_\tau b_\tau^2}{\eta_2^2 - \lambda_{\tau}},
$$
which is contradictory to our regularization assumption in \cref{theorem_FTT}.

\subsection{Proof of Lemma~\ref{lemma:FTT_error}}
This lemma requires similar techniques in \cref{lemma:bound_training_error}, which we encourage to go over first.

Using FTT, the gradients of the parameters are
\begin{align}
\partial_t \mW_{sl}(t) &= \left( \mathbb E[\vx^\top y]- \mH \vv(t) \right)\vb_{sl}(t)^\top 
\\
\partial_t \vb(t) &=  \mW(t)^\top \left( \mathbb E[\vx^\top y]- \mH\vv(t) \right).
\end{align}
Together, the gradient of $\vv(t)$ is 
\begin{align*}
    \partial \vv(t) = &= \partial_t \mW_{sl}(t) \cdot \vb_{sl}(t) + \mW(t) \cdot \partial_t \vb(t) \\
    &= \left(\mW_{ul} \mW_{ul}^\top + \mW_{sl}(t)\mW_{sl}(t)^\top + \|\vb_{sl}(t)\|_2^2 \mI \right) \left( \mathbb E[\vx^\top y] - \mH \vv(t)\right) \\
    &= \left(\mW_{ul} \mW_{ul}^\top + \mW_{sl}(t)\mW_{sl}(t)^\top + \|\vb_{sl}(t)\|_2^2 \mI \right) \mH \left( \vv_{tr}^* -  \vv(t)\right).
\end{align*}
We still denote $\mA(t) = \left(\mW_{ul} \mW_{ul}^\top + \mW_{sl}(t)\mW_{sl}(t)^\top + \|\vb_{sl}(t)\|_2^2 \mI \right) $.
Using an analysis similar to \cref{lemma:bound_training_error}, since we still have
$$
\partial_t \left[ \mW_{sl}^\top(t) \mW_{sl}(t) - \vb_{sl}(t)\vb_{sl}(t)^\top \right]= 0,
$$
we can show that $\mA(t) - (\sqrt{c_0^2+1} - 1)\mI$ is positive definite.

We next analyze the convergence of $\vv(t)$. Define the energy function (weighted error norm) as $V(t) = (\vv(t) - \vv_{tr}^*)^\top \mH (\vv(t) - \vv_{tr}^*)$.
Differentiating $V(t)$ with respect to time and substituting $\partial_t \vv(t) = \mA(t) \mH (\vv_{tr}^* - \vv(t))$:
\begin{align}
    \frac{d}{dt} V(t) &= 2 (\vv(t) - \vv_{tr}^*)^\top \mH \partial_t \vv(t) \nonumber \\
    &= 2 (\vv(t) - \vv_{tr}^*)^\top \mH \left[ \mA(t) \mH (\vv_{tr}^* - \vv(t)) \right] \nonumber \\
    &= -2 (\vv(t) - \vv_{tr}^*)^\top \mH \mA(t) \mH (\vv(t) - \vv_{tr}^*).
\end{align}
Let $\mathbf{u}(t) = \mH (\vv(t) - \vv_{tr}^*)$. The quadratic form becomes $\mathbf{u}(t)^\top \mA(t) \mathbf{u}(t)$. Since we have established that $\mA(t) \succeq (\sqrt{c_0^2+1} - 1)\mI$, let $\mu = \sqrt{c_0^2+1} - 1$, then:
\begin{align}
    \frac{d}{dt} V(t) &\leq -2 \mu \|\mathbf{u}(t)\|_2^2 \nonumber \\
    &= -2 \mu (\vv(t) - \vv_{tr}^*)^\top \mH^2 (\vv(t) - \vv_{tr}^*) \nonumber \\
    &\leq -2 \mu \lambda_{\min}(\mH) V(t),
\end{align}
where the last inequality follows from $(\vv - \vv_{tr}^*)^\top \mH^2 (\vv - \vv_{tr}^*) \ge \lambda_{\min}(\mH) (\vv - \vv_{tr}^*)^\top \mH (\vv - \vv_{tr}^*)$.
Applying Grönwall's inequality yields exponential decay for the error term $\delta(t) \triangleq \vv(t) - \vv_{tr}^*$:
\begin{align}
\label{eq:ode_bound_new}
    \|\delta(t)\|_\mH^2 = V(t) \leq V(0) \exp\left( -2 (\sqrt{c_0^2+1} - 1) \lambda_{\min}(\mH) t \right).
\end{align}

Eventually, we set
$$
\hat \vb = \scalete\mW_{FTT}(t)\vb{(t)} - \frac{1-\alpha}{c_1\alpha - (1-\alpha)} \begin{pmatrix}
            \hat  \vb_{ul} \\ \mathbf 0
        \end{pmatrix}.
$$
Since $\ell_{te}(\mW_{FTT}(t)) \leq \frac 1 2 \mathbb E _{\vx_2 = \epsilon_2} \left\|\vx \mW_{FTT}(t) \hat \vb - y \right\|^2$, we have
\begin{align*}
    \ell_{te}(\mW_{FTT}(t)) 
    &\leq \frac 1 2 \mathbb E _{\vx_2 = \epsilon_2} \left\|\vx \mW_{FTT}(t) \hat \vb - y \right\|^2 \\
    &\leq \frac 1 2 \mathbb E _{\vx_2 = \epsilon_2} \left\|\frac {c_1}{c_1 \alpha - (1-\alpha)}\vx \vv(t) - \frac{1-\alpha}{c_1\alpha - (1-\alpha)} \vx \mW_{ul}\hat \vb_{ul} - y \right\|^2 \\
    &\leq \text{err}_{te}^* + \frac 1 2 \mathbb E _{\vx_2 = \epsilon_2} \left\|\frac {c_1}{c_1 \alpha - (1-\alpha)}\vx (\vv_{tr}^* + \delta(t)) - \frac{1-\alpha}{c_1\alpha - (1-\alpha)} \vx\begin{pmatrix}
        \beta \\ c_1 \gamma
    \end{pmatrix} - \vx \beta \right\|^2 \\
    &\leq \text{err}_{te}^* + \frac 1 2 \mathbb E _{\vx_2 = \epsilon_2} \left\|\frac {c_1}{c_1 \alpha - (1-\alpha)}\vx \delta(t) \right\|^2 \\
    &\leq \text{err}_{te}^* + \mathcal O \left(\delta^\top (t) \begin{pmatrix}
        \mSigma & \mathbf 0 \\ \mathbf 0 & \eta_2^2 \mI
    \end{pmatrix} \delta(t)\right) \\
    &\leq \text{err}_{te}^* + \frac 1 2  \|\mSigma\|_2 \left( \frac {c_1}{c_1 \alpha - (1-\alpha)} \right)^2 \left\|\delta(t)\right\|^2 \\
    &\leq \text{err}_{te}^* + \frac 1 2 \|\mSigma\|_2\|\mH^{-1}\|_2 \left( \frac {c_1}{c_1 \alpha - (1-\alpha)} \right)^2 \mathcal O\left( \left\|\delta(t)\right\|^2_\mH \right) \\
    &\leq \text{err}_{te}^* + \frac 1 2 \|\mSigma\|_2\|\mH^{-1}\|_2 \left( \frac {c_1}{c_1 \alpha - (1-\alpha)} \right)^2 \mathcal O\left(\frac 1 t \right).
\end{align*}
Note that in the final step, we used the fact that the exponential decay derived in \Cref{eq:ode_bound_new} is faster than $\mathcal O(1/t)$, validating the lemma's bound.